\documentclass[10pt,twocolumn,letterpaper]{article}

\usepackage[pagenumbers]{cvpr} %

\usepackage[pagebackref,breaklinks,colorlinks,allcolors=cvprblue]{hyperref}
\definecolor{cvprblue}{rgb}{0.21,0.49,0.74}

\newcommand{\weblink}[0]{https://hosc-nn.github.io/}

\definecolor{red}{rgb}{1.0000,0.5686,0.5059}
\definecolor{orange}{rgb}{1.0000,0.7373,0.5059}
\definecolor{yellow}{rgb}{1.0000,0.8431,0.5059}

\usepackage{lipsum}
\usepackage{comment}
\usepackage{color,array}
\usepackage{xcolor}
\usepackage{colortbl}
\usepackage{graphicx}
\usepackage{amsmath}
\usepackage{subcaption}
\usepackage{float}
\usepackage{calc}     %
\usepackage{array}    %
\usepackage{multirow} %
\usepackage{balance}
\usepackage{multicol}

\def\paperID{hosc} %
\def\confName{CVPR}
\def\confYear{2026}

\title{HOSC: A Periodic Activation with Saturation Control for High-Fidelity Implicit Neural Representations}

\author{
Michal Jan Wlodarczyk\\
Warsaw University of Technology\\
{\tt\small michal.wlodarczyk@pw.edu.pl}
\and
Danzel Serrano\\
New Jersey Institute of Technology\\
{\tt\small ds867@njit.edu }
\and
Przemyslaw Musialski\\
New Jersey Institute of Technology\\
{\tt\small przem@njit.edu}
}

\begin{document}
\maketitle
\begin{abstract}
Periodic activations such as sine preserve high-frequency information in implicit neural representations (INRs) through their oscillatory structure, but often suffer from gradient instability and limited control over multi-scale behavior. We introduce the Hyperbolic Oscillator with Saturation Control (HOSC) activation, $\text{HOSC}(x) = \tanh\bigl(\beta \sin(\omega_0 x)\bigr)$, which exposes an explicit parameter $\beta$ that controls the Lipschitz bound of the activation by $\beta \omega_0$. This provides a direct mechanism to tune gradient magnitudes while retaining a periodic carrier. We provide a mathematical analysis and conduct a comprehensive empirical study across images, audio, video, NeRFs, and SDFs using standardized training protocols.  Comparative analysis against SIREN, FINER, and related methods shows where HOSC provides substantial benefits and where it achieves competitive parity. Results establish HOSC as a practical periodic activation for INR applications, with domain-specific guidance on hyperparameter selection. 
For code visit the project page:~{\small\url{\weblink}}. 

\end{abstract}

\section{Introduction}

Implicit Neural Representations (INRs) have emerged as a powerful paradigm for signal encoding, replacing discrete grids with continuous coordinate-based functions learned by neural networks~\cite{sitzmann2020siren,tancik2020ffn,mueller2022instant}. They represent a signal 
\( s : \mathbb{R}^k \to \mathbb{R}^m \) 
with a coordinate-based neural network \( f_{\theta} \) and are trained to minimize a reconstruction loss (plus regularization) over dense coordinate samples. Their promise—infinite resolution, memory efficiency, and natural handling of irregular geometries—has driven rapid adoption across audio~\cite{kun2022INRAS}, video~\cite{chen2021nerv}, image compression~\cite{dupont2021coin}, and 3D reconstruction~\cite{mildenhall2020nerf}.  

\begin{figure}[t]
    \centering    \includegraphics[width=0.49\textwidth]{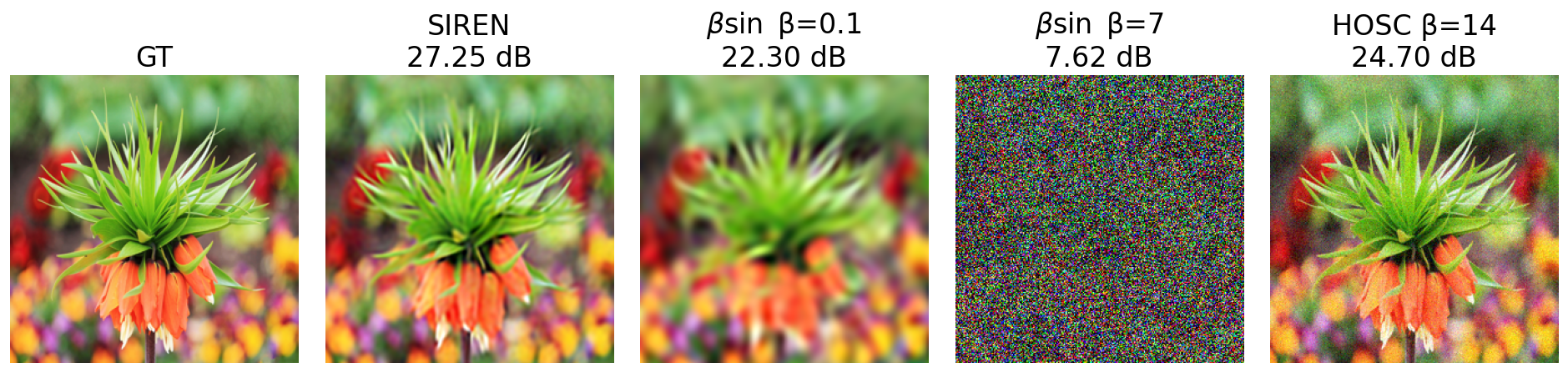}
    \includegraphics[width=0.49\textwidth]{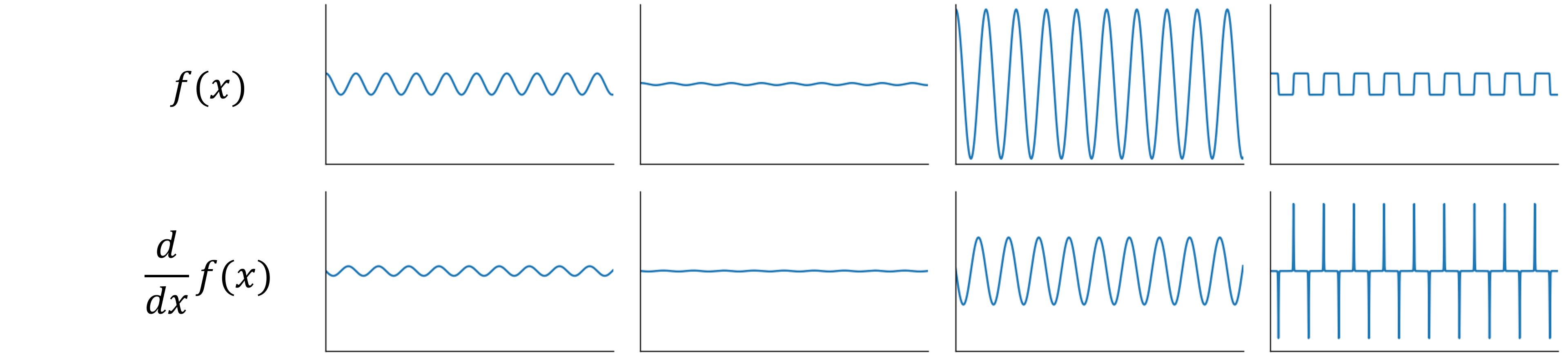}
    \caption{
    2D image fitting with an identical coordinate MLP and fixed frequency \(\omega_0 = 30\), varying the activation.
    From left to right: ground truth, SIREN \(\sin(\omega_0 x)\), scaled sine \(\beta\sin(\omega_0 x)\), and HOSC \(\tanh(\beta\sin(\omega_0 z))\). 
    SIREN and low-\(\beta\) sine underfit high-frequency detail, large-\(\beta\) sine produces unstable artifacts, while HOSC recovers sharp structure by combining high-frequency support with saturated, Lipschitz-controlled gradients.
    }
    \label{fig:teaser}
\end{figure}

Despite this success, INRs face a fundamental limitation: learning high-frequency components is inherently difficult for standard multilayer perceptrons, a phenomenon known as spectral bias~\cite{rahaman2019spectral,basri2020frequency,li2025inrbench}. Two main strategies have emerged to mitigate this: positional encoding methods~\cite{tancik2020ffn,mildenhall2020nerf} that transform input coordinates to enhance high-frequency learning, and periodic activation functions~\cite{sitzmann2020siren} that directly improve representational capacity for oscillatory structure.

Within periodic activation designs, however, a fundamental architectural limitation persists. Fixed-period sine activations $\sin(\omega_0 x)$, as in SIREN~\cite{sitzmann2020siren}, have activation-level Lipschitz constant \(L_{\text{SIREN}}=\omega_0\), so the same parameter \(\omega_0\) simultaneously controls spectral support and gradient scale. Variable-periodic approaches such as FINER~\cite{liu2024finer} address frequency flexibility through bias initialization and amplitude-dependent period, but do not provide a simple global Lipschitz parameter: gradient behaviour remains inherited from the carrier frequency and internal modulation. 
As highlighted by recent benchmarking~\cite{li2025inrbench}, the absence of an explicit, interpretable gradient-scale knob in periodic activations limits multi-modal adaptivity.

To address it, we introduce a \emph{Hyperbolic Oscillator with Saturation Control} (HOSC), a periodic activation function with saturation control that enables explicit gradient tuning through a compositional design, defined as: %
\[
\mathrm{HOSC}_\beta(x) = \tanh\!\big(\beta \sin(\omega_0 x)\big),
\]
where \(\omega_0\) sets the carrier frequency and \(\beta>0\) controls saturation strength. This composition preserves periodic structure while inducing a tight activation-level Lipschitz constant \(L_{\text{HOSC}} = \beta \omega_0\). In practice, \(\omega_0\) can be chosen for the desired spectral support as in SIREN, and \(\beta\) then acts as a single, interpretable parameter that multiplicatively scales gradients. HOSC thus provides a simple activation-level mechanism for per-modality gradient adaptation, without changing encodings or architecture, and with bounded outputs for numerical stability. 

Figure~\ref{fig:teaser} illustrates these behaviours on a 2D image: for a fixed network and carrier frequency, SIREN and low-\(\beta\) scaled sine yield oversmoothed reconstructions, a high-\(\beta\) scaled sine becomes unstable, whereas HOSC attains a sharp reconstruction by combining the same spectral support with controlled, saturated gradients.
In summary, our contributions are:
\begin{itemize}
\item We propose HOSC, a periodic activation with saturation control that introduces an explicit activation-level Lipschitz parameter \(L_{\text{HOSC}} = \beta \omega_0\), separating frequency selection (\(\omega_0\)) from gradient scaling (\(\beta\)). In controlled ablations, we show that HOSC remains stable and improves reconstruction quality in \(\beta\)-regimes where scaled sine activations \(\beta\sin(\omega_0 x)\) collapse. 
\item We provide a drop-in replacement for SIREN that requires no changes to positional encodings, remains compatible with standard SIREN-style initialization, and maintains bounded outputs for stable optimization.
\item We validate HOSC across multiple modalities demonstrating 
significant performance gain on audio reconstruction and measurable improvements for video and images over prior periodic activations (SIREN, FINER), while providing competitive results on NeRFs and SDFs.
\end{itemize}

\section{Related Work}\label{sec:related}

\subsection{INRs, Spectral Bias, and Periodic Activations}

Implicit neural representations (INRs) parameterize continuous signals via coordinate MLPs and have been applied to images, videos, audio, and 3D geometry~\cite{sitzmann2020siren,mildenhall2020nerf,tancik2020ffn}. A central challenge is \emph{spectral bias}: standard MLPs fit low frequencies first~\cite{rahaman2019spectral,basri2020frequency,cao2019towards}, which NTK theory explains through eigenvalue decay of the kernel~\cite{jacot2018neural}. Fourier features~\cite{tancik2020ffn} mitigate this by mapping coordinates through random Fourier embeddings that reshape the NTK into a stationary kernel with tunable bandwidth, thereby modifying the \emph{frequency support} of the model.

SIREN~\cite{sitzmann2020siren} addresses spectral bias from the activation side by using fixed-period sine activations $\sin(\omega_0 z)$, with activation-level Lipschitz constant $L_{\text{act}} = \omega_0$ and bounded outputs. 
This improves representation of derivatives and PDE solutions, but the Lipschitz constant is entirely determined by $\omega_0$, so frequency support and gradient scale are coupled. Increasing $\omega_0$ simultaneously increases the highest representable frequency and the upper bound on activation derivatives, leaving no separate control over gradient magnitudes during training.

\subsection{Variable-Periodic, Gaussian, and Learned}

Several works extend SIREN along the \emph{frequency axis}. WIRE~\cite{saragadam2023wire} introduces complex Gabor activations with superior joint space–frequency localization at the price of complex arithmetic and higher parameter counts. FINER~\cite{liu2024finer} and FINER++~\cite{zhu2024finerpp} propose variable-periodic activations of the form $\sin(\omega_0 (|\mathbf{y}|+1)\mathbf{y})$ and show, via geometric and NTK analysis, that widening bias initialization ranges $b\sim\mathcal{U}(-k,k)$ lets the network select sub-functions with different effective frequencies. Conceptually, FINER tunes the \emph{supported frequency set} through a combination of variable period and bias range $k$, but does not introduce an explicit global gradient or Lipschitz parameter: derivatives grow with $|z|$ and are not bounded by a single scalar. FINER++ extends this variable-periodic idea to Gaussian and wavelet carriers.

Orthogonal branches replace periodicity altogether. Gaussian activations~\cite{ramasinghe2022periodicity} were shown to outperform sine on several INR tasks, and GARF~\cite{chng2022garf} adapts Gaussian activations to NeRF. These works emphasize localization and gradient preservation rather than periodic structure. Sinc and related activations have also been studied from sampling-theoretic and functional-analytic perspectives~\cite{saratchandran2024sampling,meronen2021periodic}, but numerical issues near $x=0$ limit practical adoption. More recently, meta-learned activation dictionaries (MIRE)~\cite{jayasundara2025mire} and functional architectures such as KAN~\cite{liu2024kan} treat activation choice as a learning problem, confirming that activation geometry (including its Lipschitz and smoothness properties) strongly influences INR performance, but at the cost of additional infrastructure and complexity.

INR-Bench~\cite{li2025inrbench} systematically evaluates 56 coordinate-MLP variants and 22 KAN models across nine INR tasks, providing L-Lipschitz and L-smooth constants for common activations (e.g., $L_{\text{Lip}}=\omega$ for sine, $(1/\sigma)e^{-1/2}$ for Gaussians). Its main conclusion is that periodic sine combines favorable high-frequency learning in low-dimensional domains with a \emph{fixed} activation Lipschitz constant, which limits multi-scale adaptivity across modalities. This motivates activations with explicit, interpretable control over Lipschitz and gradient scale.

\subsection{Architectural and Encoding Approaches}

A different line of work targets multi-scale representation via \emph{architecture} or \emph{encoding}, rather than activation shape. Multiplicative Filter Networks and Residual MFN~\cite{fathony2021mfn,shekarforoush2022residualmfn} use multiplicative filters and skip connections to represent multiple scales, and ACORN and MINER~\cite{lindell2021acorn,saragadam2022miner} decompose signals spatially or through mixture-of-experts to fit gigapixel images. InstantNGP~\cite{mueller2022instant} and Mip-NeRF variants~\cite{barron2021mipnerf,barron2022mipnerf360} show that multi-resolution hash grids and anti-aliasing encodings can dramatically accelerate NeRF and mitigate high-frequency artifacts. BACON~\cite{lindell2022bacon} and anti-aliased SDF methods~\cite{zhuang2023antialias} achieve band-limited or LOD-controlled representations via encoding-level constraints, while NGSSF~\cite{mujkanovic2024ngssf} uses Lipschitz-constrained MLPs together with modulated encodings to realize Gaussian scale spaces. Explicit-primitive approaches such as Neural Point Catacaustics~\cite{kopanas2022catacaustics} bypass INRs for certain phenomena (e.g., caustics) and thereby sidestep some low-frequency biases.

These methods primarily act along the \emph{architectural/encoding} axis: they change how inputs are mapped and combined, not the local activation Lipschitz of a simple coordinate MLP. HOSC is designed to be orthogonal and compatible with such approaches: it modifies the activation while leaving encodings and architectures unchanged.

\section{Method --- HOSC}\label{sec:method}

\begin{figure}[h]
    \centering   
    \includegraphics[width=0.48\textwidth]{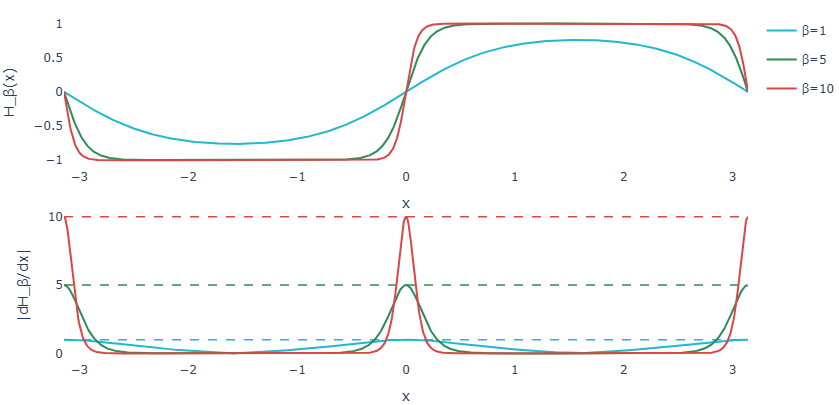}
    \caption{{HOSC gradient control.} 
Wrapping the sine in a hyperbolic tangent results in a periodic activation function that achieves saturation control through $\beta$. 
Derivative magnitude with Lipschitz bounds $L=\beta\omega_0$ (dashed). 
Increasing $\beta$ amplifies and localizes gradients at zero-crossings while 
maintaining bounded outputs.}

    \label{fig:plot_hosc}
\end{figure}

\subsection{Hyperbolic Oscillator with Saturation Control}

Our goal is to introduce a mechanism to decouple the gradient magnitude for different signal modalities or training regimes of a single periodic activation without sacrificing the periodic structure. Wrapping the sine carrier in a hyperbolic tangent results in an oscillating activation function that achieves saturation control trough a hyperbolic tangent: 
\[
\text{HOSC}_\beta(x) = \tanh(\beta \sin(\omega_0 x))
\]
where \(\beta > 0\) controls saturation strength and \(\omega_0\) determines carrier frequency. This composition introduces two key properties: (1) bounded outputs \(|\text{HOSC}_\beta(x)| \leq 1\), and (2) tunable gradient magnitude through the \(\beta\) parameter.

Figure~\ref{fig:plot_hosc} visualizes HOSC for varying \(\beta\) values. At low \(\beta < 1\), the tanh saturation is weak, producing near-linear behavior around zero and compressed oscillations at extrema. At high \(\beta > 5\), saturation becomes aggressive, approximating a square wave with sharp transitions. This behavior enables HOSC to interpolate between smooth periodic functions (low \(\beta\)) and step-like responses (high \(\beta\)), providing flexibility across signal types requiring different frequency-detail tradeoffs.

\subsection{Gradient Analysis and Lipschitz Bounds}

To understand HOSC's gradient flow properties, we compute its first derivative using the chain rule. For \(\text{HOSC}_\beta(x) = \tanh(\beta \sin(\omega_0 x))\), we obtain:
\[
\frac{d}{dx}\text{HOSC}_\beta(x) = \beta \omega_0 \cos(\omega_0 x) \cdot \operatorname{sech}^2(\beta \sin(\omega_0 x))
\]
where \(\operatorname{sech}^2(z) = 1/\cosh^2(z) = 1 - \tanh^2(z)\). This derivative exhibits three key characteristics: (1) multiplicative scaling by \(\beta \omega_0\), (2) periodic oscillation via \(\cos(\omega_0 x)\), and (3) amplitude modulation through \(\operatorname{sech}^2(\cdot)\) that depends on the sine carrier's current value.

\noindent
\textbf{Activation-level Lipschitz constant.} 
Since \(|\cos(\omega_0 x)| \leq 1\) for all \(x\) and \(0 < \operatorname{sech}^2(z) \leq 1\) for all \(z \in \mathbb{R}\) (with equality at \(z = 0\)), we obtain a tight Lipschitz bound:
\[
\left|\frac{d}{dx}\text{HOSC}_\beta(x)\right| \leq \beta \omega_0
\]
This bound is achieved precisely when \(\sin(\omega_0 x) = 0\) (making \(\operatorname{sech}^2(0) = 1\)) and \(\cos(\omega_0 x) = \pm 1\), which occurs at \(x = k\pi/\omega_0\) for integer \(k\). The activation-level Lipschitz constant \(L = \beta \omega_0\) bounds the derivative magnitude at each nonlinearity during backpropagation.

\subsection{Layer and Network Gradient Bounds}

Having established the scalar activation-level Lipschitz bound \(L_{\text{HOSC}} = \beta \omega_0\) for the activation function, we now examine how this property propagates through multi-layer networks—a critical step for understanding HOSC's impact on training dynamics.

\noindent
\textbf{From activation to layer.} For a standard MLP layer computing \(\mathbf{y} = \text{HOSC}_\beta(\mathbf{W}\mathbf{x} + \mathbf{b})\), the chain rule decomposes the gradient into activation and linear components. Since the activation-level Jacobian satisfies \(\|\nabla_{\mathbf{z}} \text{HOSC}_\beta(\mathbf{z})\|_2 \leq \beta |\omega_0|\), the full layer gradient is bounded by
\[
\|\nabla_{\mathbf{x}} \mathbf{y}\|_2 \leq \beta |\omega_0| \, \|\mathbf{W}\|_2
\]
where \(\|\mathbf{W}\|_2\) is the spectral norm of the weight matrix. This shows that gradient scale at each layer is jointly controlled by the activation parameter \(\beta\) and weight magnitude—enabling practitioners to adjust \(\beta\) to compensate for large weights without retraining.

\noindent
\textbf{Deep network composition.} Stacking \(L\) such layers, the network-level Lipschitz constant satisfies the multiplicative upper bound
\[
L_{\text{net}} \leq \prod_{\ell=1}^L \beta |\omega_0| \, \|\mathbf{W}_\ell\|_2.
\]
This bound reveals why \(\beta\) acts as a global gradient scaling knob: reducing \(\beta\) by half reduces this upper bound on the network Lipschitz constant by a factor of \(2^{-L}\), providing exponential control over gradient magnitudes in deep architectures.

\subsection{Asymptotic Behavior and Practical Regimes}\label{sec:asymp_behavior}

To understand HOSC's behavior across the full \(\beta\)-spectrum—from smooth near-linear responses to sharp square-wave transitions—we analyze limiting cases and relate them to the regimes that empirically arise in our experiments.

\noindent
\textbf{Smooth regime (\(\beta \to 0\)).}
For small \(\beta\), a Taylor expansion of the outer \(\tanh\) yields
\(\mathrm{HOSC}_\beta(x) \approx \beta \sin(\omega_0 x)\), recovering a gently scaled sine with small derivative magnitude. In this regime, gradients remain smooth and relatively uniform, which empirically aligns with the behaviour we observe on high-dimensional coordinate tasks such as video and NeRF, where smaller \(\beta\) values lead to stable optimization (Sec.~4).

\noindent
\textbf{Sharp regime (\(\beta \to \infty\)).}
Conversely, as \(\beta\) grows large, the activation approaches the square-wave limit
\(\mathrm{HOSC}_\beta(x) \to \mathrm{sign}(\sin(\omega_0 x))\), and the derivative concentrates in narrow neighborhoods around the zeros of \(\sin(\omega_0 x)\). This produces highly localized, high-magnitude gradients that are well suited to fitting sharp transitions. In our one-dimensional audio experiments, moderate-to-large \(\beta\) values exploit this effect and substantially improve reconstruction of rapid waveform variations (Sec.~4).

\noindent
\textbf{Gating width.}
The derivative
\(\mathrm{HOSC}_\beta'(x) = \beta \omega_0 \cos(\omega_0 x)\,\mathrm{sech}^2(\beta \sin(\omega_0 x))\)
contains a \(\mathrm{sech}^2(\beta \sin(\cdot))\) factor that compresses gradients away from the zeros of the sine carrier. For thresholds proportional to the peak derivative, e.g.\ \(\tau = \kappa \beta \omega_0\) with fixed \(\kappa \in (0,1)\), one can show (cf. Appendix~\ref{sec:appx:asymptotics}) that the measure of the superlevel set
\(\{x : |\mathrm{HOSC}_\beta'(x)| \geq \tau\}\) scales on the order of \(1/\beta\). Thus increasing \(\beta\) narrows the region where gradients are large while simultaneously raising their maximum magnitude, providing a controllable trade-off between gradient strength and spatial coverage that we exploit differently across modalities in Sec.~4.

\noindent
\textbf{Mathematical properties.}
For completeness, we note that HOSC inherits the sine carrier's \(2\pi/\omega_0\)-periodicity and odd symmetry, while adding bounded outputs \(\mathrm{HOSC}_\beta(x) \in [-1,1]\) and \(C^\infty\)-smoothness. The activation-level Lipschitz bound \(L_{\text{act}} = \beta \omega_0\) is tight: at \(x = k\pi/\omega_0\), for integer \(k\), we have \(\sin(\omega_0 x)=0\), \(\mathrm{sech}^2(0)=1\), and \(|\mathrm{HOSC}_\beta'(x)| = \beta \omega_0\).

\subsection{NTK view of HOSC-$\beta$}
\label{subsec:ntk-analysis}

To complement the Lipschitz analysis, we inspect the empirical NTK in a
simple one–hidden-layer INR with shared random parameters. On a 1D grid
\(\{x_i\}\) we compute \(K_\theta(x_i,x_j)\) for SIREN and for HOSC at
different \(\beta\), averaging over several seeds.

\begin{figure}[t]
    \centering
    \includegraphics[width=0.99\linewidth]{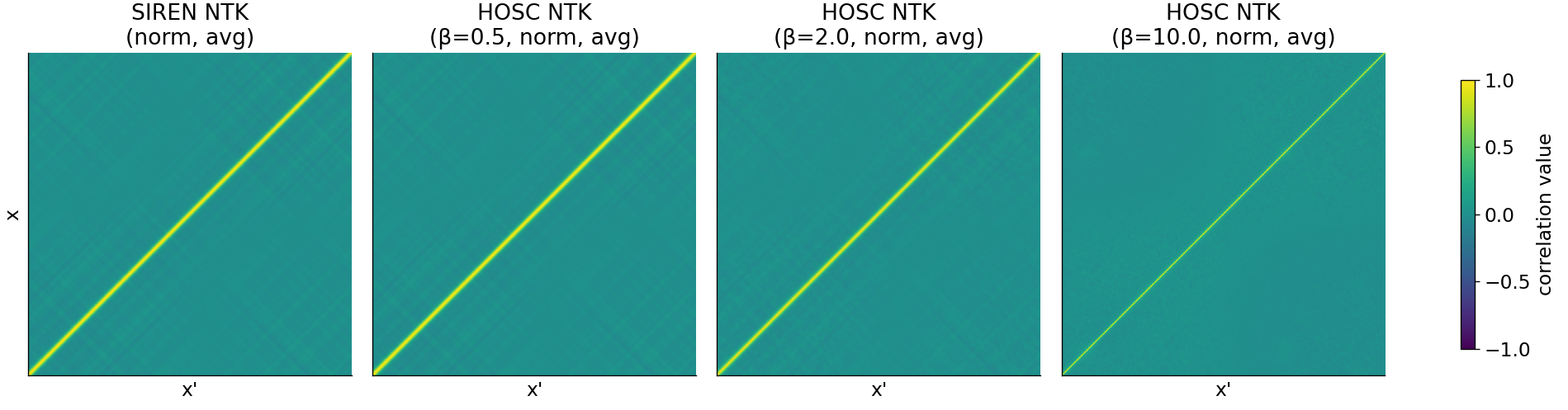}\\
    \includegraphics[width=0.99\linewidth]{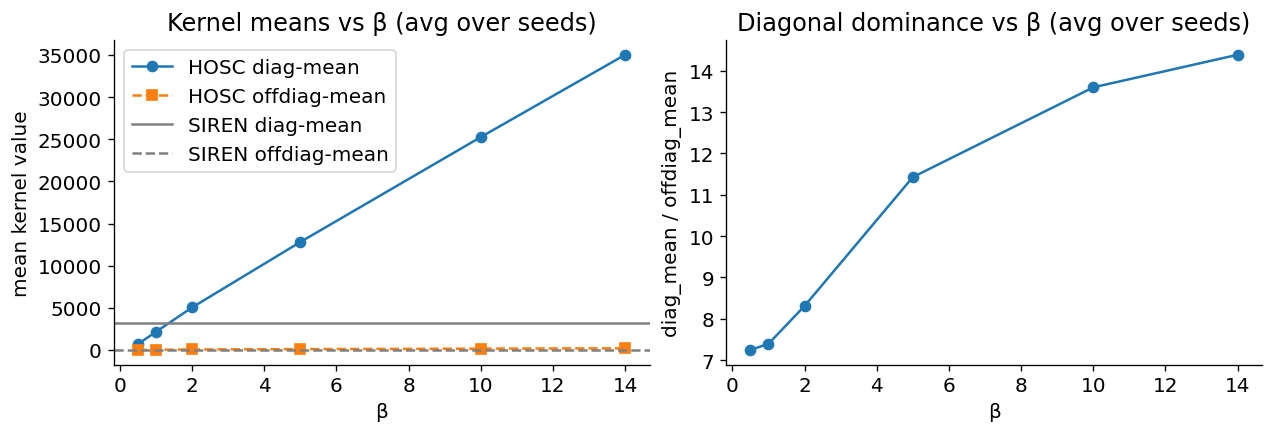}
    \caption{
        {Empirical NTK behaviour of HOSC.}
        \emph{Top:} Correlation-normalized kernels for SIREN and HOSC with
        $\beta\in\{0.5,2,10\}$, averaged over multiple seeds.  Larger $\beta$ sharpens the diagonal and attenuates off–diagonal structure. 
        \emph{Bottom:} Mean diagonal values and diagonal-dominance ratio
        (diag / off-diagonal std) as a function of $\beta$.  Both increase
        steadily, indicating that $\beta$ acts as a controllable kernel-scale and interaction-strength knob.
    }
    \label{fig:hosc-ntk}
\end{figure}

Because \(\beta\) rescales \(K_\theta\), we normalize by the diagonal and
examine the correlation kernel
\(\tilde K_{ij}=K_{ij}/\sqrt{K_{ii}K_{jj}}.\)
Figure~\ref{fig:hosc-ntk} (top) shows that for small \(\beta\) the HOSC NTK
is nearly indistinguishable from SIREN, while larger \(\beta\) slightly
sharpens the diagonal band and attenuates off–diagonal correlations.

We summarize this behaviour with two scalar statistics
(Fig.~\ref{fig:hosc-ntk}, bottom): the mean diagonal of \(K_\theta\)
increases almost linearly with \(\beta\), and a diagonal-dominance measure
(diagonal mean divided by off–diagonal standard deviation) grows
monotonically. In this one-layer setting, \(\beta\) therefore acts as a
controllable kernel-scale and interaction-strength knob, smoothly shifting
the kernel from a SIREN-like regime at small \(\beta\) toward a more
self-dominated regime at larger \(\beta\).

\subsection{Relationship to SIREN}
\label{subsec:hosc-vs-siren}

A natural question is whether HOSC provides capabilities beyond standard
sinusoidal activations, or in other words: \emph{why HOSC cannot be reduced to a rescaled or reparameterized SIREN?} The fundamental difference lies in parameter
coupling: SIREN's single parameter $\omega_0$ simultaneously determines both
the spectral support (which frequencies the network can represent) and the
gradient magnitude during backpropagation. 

This coupling means that
increasing $\omega_0$ to capture higher frequencies necessarily amplifies
gradients, limiting adaptability. HOSC breaks this coupling through its
compositional design $\tanh(\beta\sin(\omega_0 x))$, where $\omega_0$ sets
the carrier frequency while $\beta$ provides an independent multiplicative
gradient control. This decoupling enables stable training in high-gradient
regimes ($\beta>5$) that cause instability in amplitude-scaled sine networks.
We verify this distinction through two controlled ablations.

\begin{figure}[t]
  \centering
  \includegraphics[width=\linewidth]{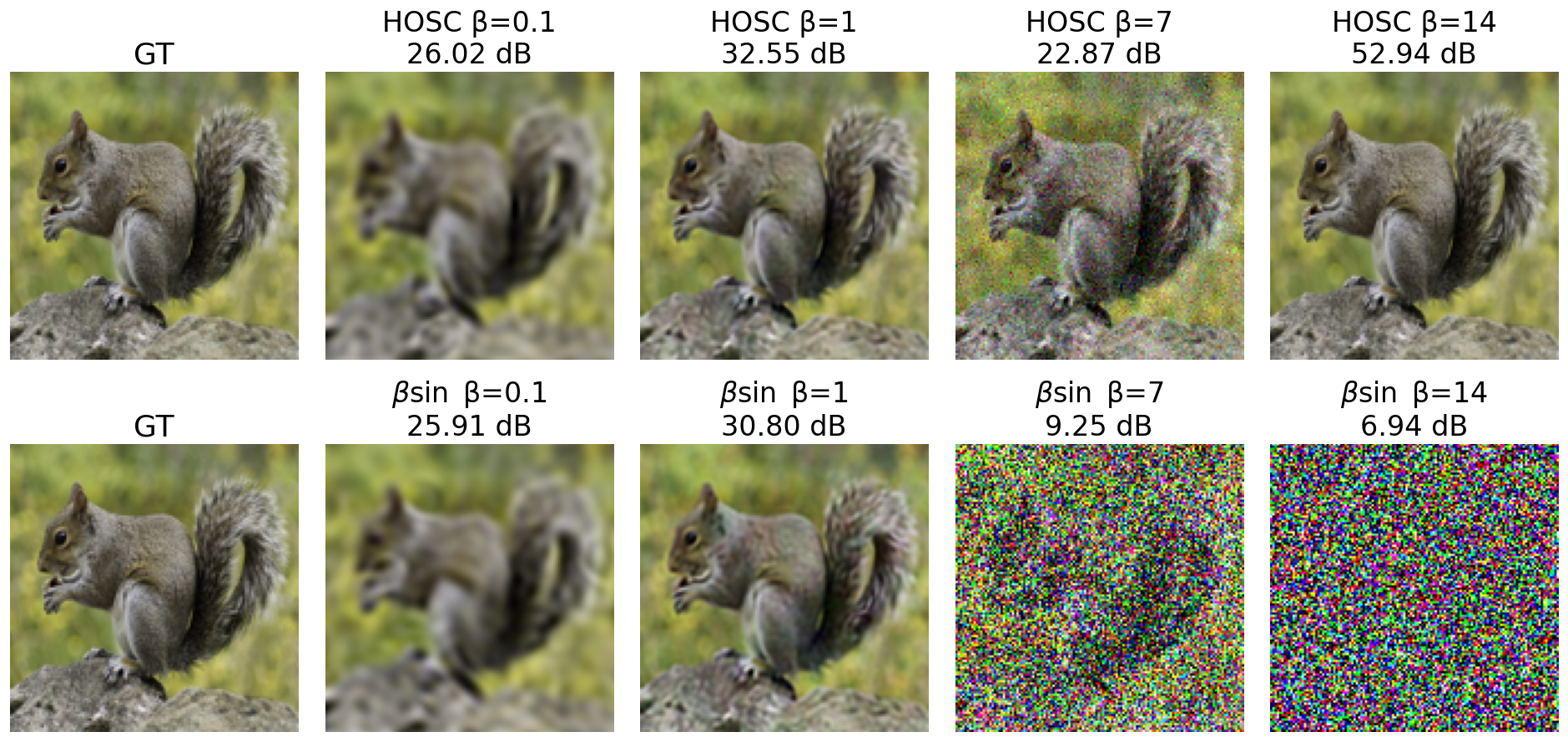}
  \caption{{Effect of $\beta$ for HOSC and $\beta$-sin.}
  Single DIV2K image, $\omega_0=30$. Top row: GT and HOSC for $\beta\in\{0.1,1,7,14\}$.
  Bottom row: GT and $\beta$-sin for the same $\beta$. All runs use the same INR architecture and training setup.}
  \label{fig:beta-sweep}
\end{figure}

\noindent
\textbf{Ablation 1: amplitude scaling (\texorpdfstring{$\beta$}{beta}-sin control).}
We test whether the tanh envelope is essential by comparing HOSC against a 
direct amplitude-scaled sine $\beta\sin(\omega_0 x)$ under identical training 
conditions (Fig.~\ref{fig:beta-sweep}). While both behave similarly at small 
$\beta$, the unscaled sine rapidly destabilizes as $\beta$ increases, whereas 
HOSC remains stable. This confirms that the 
saturating envelope—not mere amplitude scaling—enables trainability in 
high-gradient regimes.

\begin{figure}[t]
  \centering
  \includegraphics[width=\linewidth]{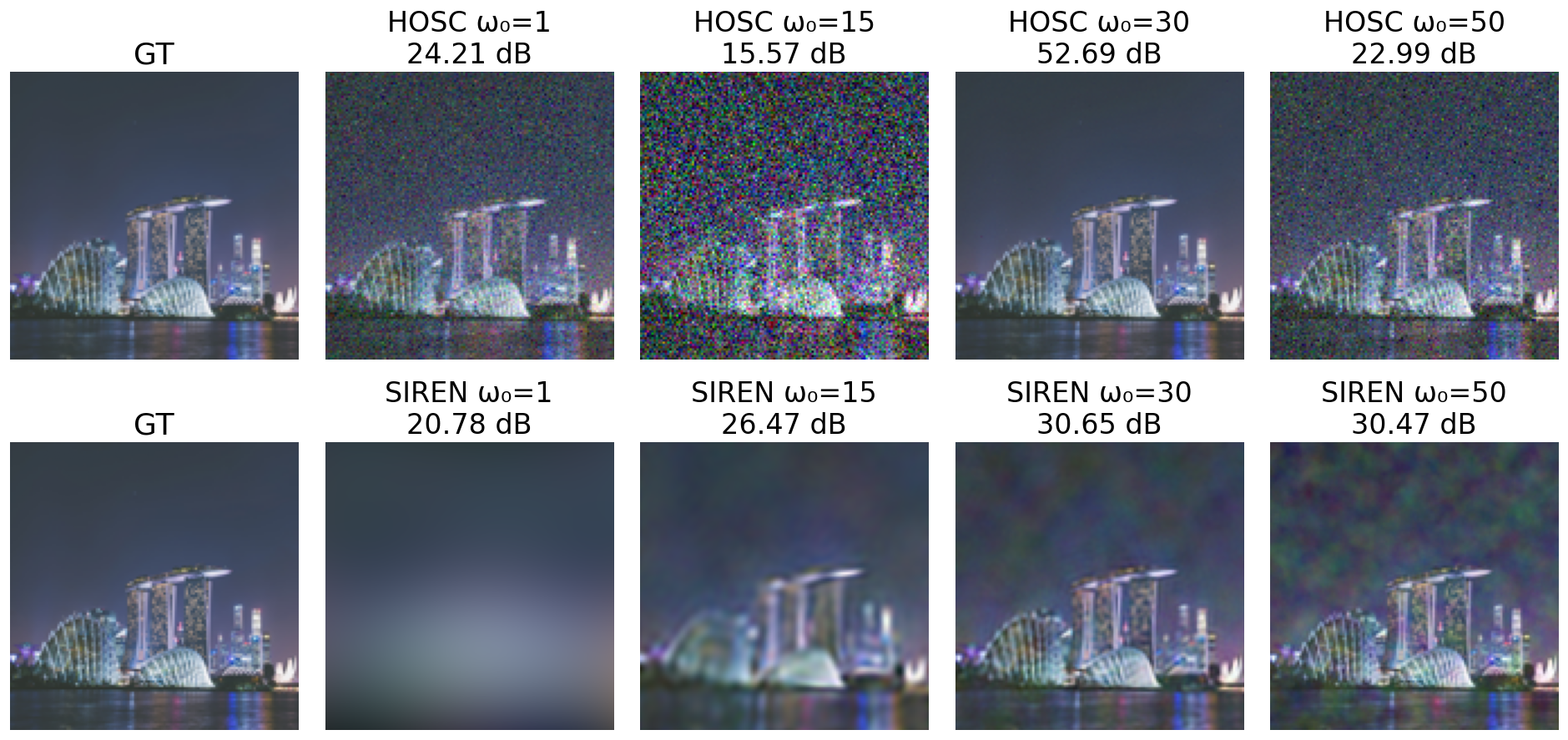}
  \caption{{Effect of $\omega_0$ for HOSC and SIREN.}
  Single DIV2K image. Top row: HOSC with fixed $\beta=14$ and $\omega_0\in\{1,15,30,50\}$.
  Bottom row: SIREN with $\beta=1$ and the same $\omega_0$ values, using the same INR architecture and training setup.}
  \label{fig:omega-sweep}
\end{figure}

\noindent
\textbf{Ablation 2: frequency tuning (\texorpdfstring{$\beta$}{beta}–vs–\texorpdfstring{$\omega_0$}{omega}).}
We test whether SIREN can match HOSC through aggressive frequency tuning by 
varying $\beta$ for HOSC and $\omega_0$ for SIREN over matched ranges 
(Fig.~\ref{fig:omega-sweep}). While both show performance peaks, HOSC 
consistently achieves higher reconstruction quality, demonstrating that 
$\beta$'s gradient control cannot be emulated through frequency 
reparameterization alone.

These ablations confirm: (1)~HOSC is not a $\beta$-scaled sine, and
(2)~HOSC is not equivalent to frequency-tuned SIREN. The tanh–sine
composition provides a distinct gradient-control mechanism that neither
amplitude scaling nor frequency reparameterization can replicate.

\subsection{Relationship to FINER}

FINER~\cite{liu2024finer} addresses spectral bias from a different axis than
HOSC. Rather than modifying the Lipschitz behaviour of a fixed-frequency
sine, FINER introduces a \emph{variable-periodic} activation of the form
\begin{equation}
    \sigma_{\text{FINER}}(x)
    = \sin\big(\omega_0 \alpha(x)\, x\big),
    \quad \alpha(x) = |x| + 1,
\end{equation}
and tunes the supported frequency set by widening the bias initialization
\(b \sim \mathcal{U}(-k,k)\). Geometric and NTK analyses in~\cite{liu2024finer}
show that increasing \(k\) selects sub-functions with higher effective
frequencies, enlarges the supported frequency set \(F_{\omega_0,k}\), and
enhances NTK diagonal dominance, thereby improving high-frequency
convergence.

In contrast, HOSC keeps the carrier frequency fixed and instead wraps the
sine with a saturating nonlinearity
\begin{equation}
    \sigma_{\text{HOSC}}(x)
    = \tanh\!\big(\beta \sin(\omega_0 x)\big),
\end{equation}
yielding a tight activation-level Lipschitz constant
\(\|\sigma'_{\text{HOSC}}\|_\infty = \beta \omega_0\) (Sec.~\ref{sec:method}).
Thus HOSC introduces an explicit, global gradient-scale parameter \(\beta\)
while leaving spectral support determined by \(\omega_0\) and the linear
layers. FINER can be viewed as an \emph{initialization-level frequency
tuner}: it enlarges and redistributes the frequency set via bias range \(k\)
and variable period \(\alpha(x)\). HOSC is an \emph{activation-level
gradient tuner}: it provides bounded outputs and a controllable Lipschitz
constant without altering the underlying frequency set. The two mechanisms
are therefore orthogonal and, in principle, composable.

\begin{figure*}[htbp]
    \centering
    
    \newlength{\SpaceCaptionToRow}
    \setlength{\SpaceCaptionToRow}{-0.0em} %
    
    \newlength{\SpaceMetricToNextRow}
    \setlength{\SpaceMetricToNextRow}{-0.2em} %
    
    \newlength{\SpaceImageToMetric}
    \setlength{\SpaceImageToMetric}{-1.3em} %

    \vspace{\SpaceCaptionToRow} 

    \def\subfigwidth{0.142\textwidth}

    \begin{subfigure}[t]{\subfigwidth}
        \centering
        {PEMLP} %
        \includegraphics[width=\textwidth]{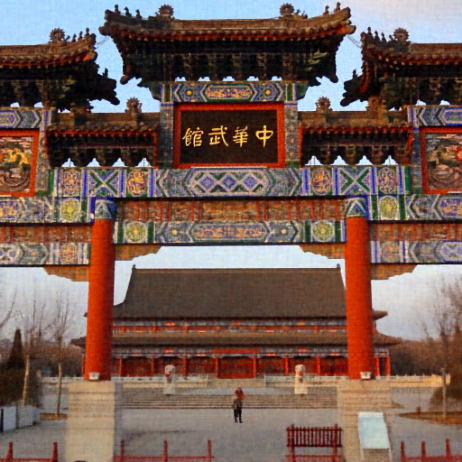}
        \vspace{\SpaceImageToMetric} %
        \caption*{{26.3002dB}} %
    \end{subfigure}%
    \hfill
    \begin{subfigure}[t]{\subfigwidth}
        \centering
        {Gauss} %
        \includegraphics[width=\textwidth]{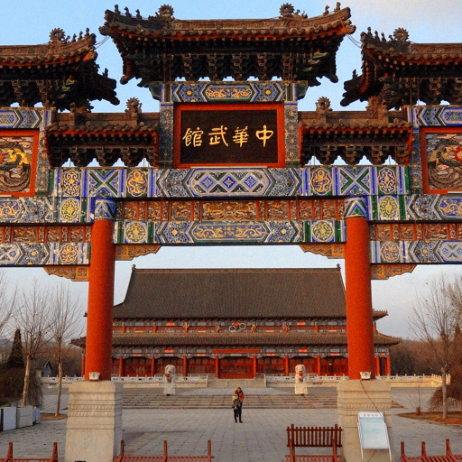}
        \vspace{\SpaceImageToMetric} %
        \caption*{{31.4108dB}} %
    \end{subfigure}%
    \hfill
    \begin{subfigure}[t]{\subfigwidth}
        \centering
        {WIRE} %
        \includegraphics[width=\textwidth]{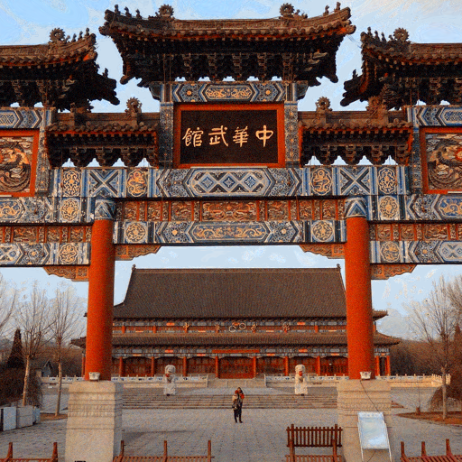}
        \vspace{\SpaceImageToMetric} %
        \caption*{{28.0027dB}} %
    \end{subfigure}%
    \hfill
    \begin{subfigure}[t]{\subfigwidth}
        \centering
        {SIREN} %
        \includegraphics[width=\textwidth]{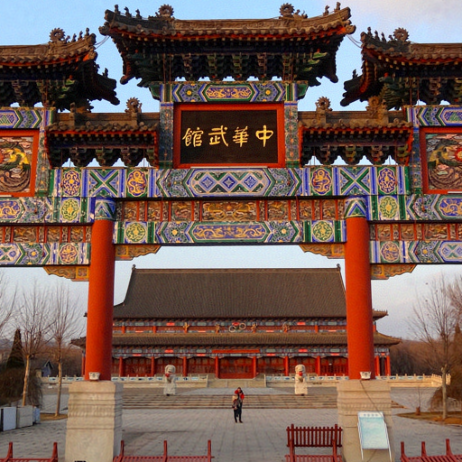}
        \vspace{\SpaceImageToMetric} %
        \caption*{{33.6837dB}} %
    \end{subfigure}%
    \hfill
    \begin{subfigure}[t]{\subfigwidth}
        \centering
        {FINER} %
        \includegraphics[width=\textwidth]{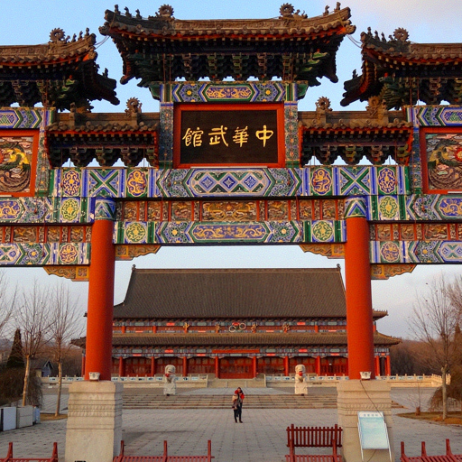}
        \vspace{\SpaceImageToMetric} %
        \caption*{{35.8982dB}} %
    \end{subfigure}%
    \hfill
    \begin{subfigure}[t]{\subfigwidth}
        \centering
        {HOSC} %
        \includegraphics[width=\textwidth]{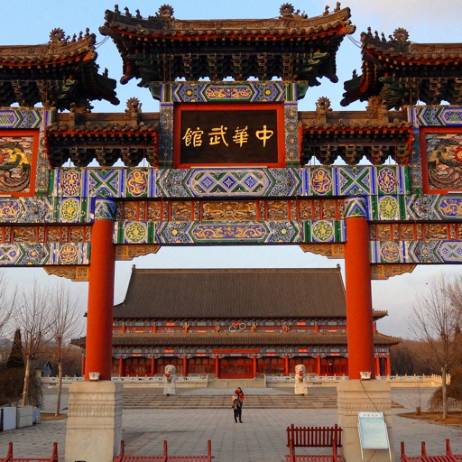}
        \vspace{\SpaceImageToMetric} %
        \caption*{\textbf{36.6901dB}} %
    \end{subfigure}%
    \hfill
    \begin{subfigure}[t]{\subfigwidth}
        \centering
        {GT} %
        \includegraphics[width=\textwidth]{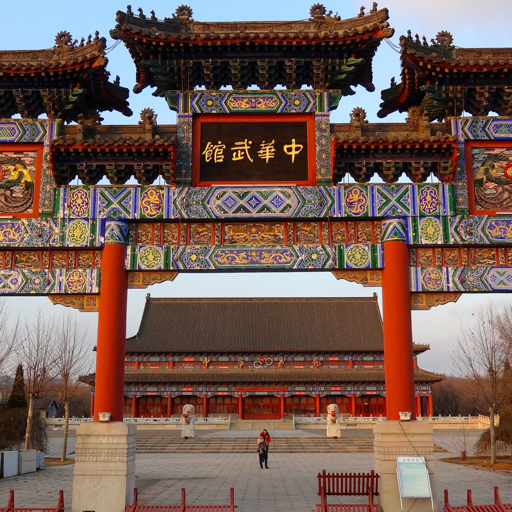} %
        \vspace{\SpaceImageToMetric} %
        \caption*{} %
    \end{subfigure}%
    \hfill

    \vspace{-0.8em}
    \caption{Qualitative comparison on image fitting. Each row reconstructs the GT data with column specified method. Quantitative PSNR results showcased per example below reconstructed images. HOSC parameter $\beta=14.0$.}
    \label{fig:image_comparison_paper}

\end{figure*}

\section{Practical Experiments}\label{sec:experiment}

We examine HOSCs reconstruction quality on taks such as image, audio, video, NeRF and SDF reconstruction. In each experiment we first explore $\beta$ parameter space with an ablation study. The best performing value is then used during method comparison.
We compare HOSC with four activation based INR methods, \textit{i.e.} Gaussian (Gauss) \cite{ramasinghe2022periodicity}, Wavelet (WIRE) \cite{saragadam2023wire}, Sine (SIREN) \cite{sitzmann2020siren} and FINER \cite{liu2024finer} activations along with encoding based methods, \textit{i.e.}, Fourier Features positional encoding (PEMLP) \cite{tancik2020ffn}. %
We report most notable results in the paper and present the details in the Appx. 
All experiments use a unified protocol with SIREN-style initialization, Adam, identical metrics, fixed seeds, deterministic dataloading, and consistent training schedules on A100/H100 GPUs.

\subsection{Image Fitting}
On 2D image reconstruction we aim to learn a mapping of 2D input pixel locations to 3D RGB output colors $f\colon\mathbb{R}^{2} \rightarrow \mathbb{R}^{3}$, the loss function is the mean $L_2$ distance between the network output and
the ground truth (MSE). We evaluate the reconstruction performance of the INR on 16 images from the DIV2k dataset \cite{Agustsson_2017_CVPR_Workshops} of $512\times 512$ resolution. Firstly we ablate the $\beta$ on all samples. We utilize the, on average, best performing $\beta$ during the comparison with SoTA methods. 

\begin{table}[bp]
\centering
\scriptsize
\setlength\tabcolsep{2pt}
\setlength{\tabcolsep}{1.8pt}
\caption{\textbf{Image HOSC $\beta$-ablation}. Averaged over 16 samples. In experiment tables we color code each cell as  \colorbox{red}{best}, \colorbox{orange}{second best}, and \colorbox{yellow}{third best}.}
\vspace{-1.2em}
\begin{tabular}{ccccccccccccccccc}
\toprule
& \multicolumn{7}{c}{\textbf{HOSC} $\boldsymbol{\beta}$} \\
\cmidrule(lr){2-9}
Metric & 
\multirow{2}*{0.1} & \multirow{2}*{0.2} &
\multirow{2}*{0.5} & \multirow{2}*{0.8} &
\multirow{2}*{1.0} & \multirow{2}*{2.0} &
\multirow{2}*{4.0} & \multirow{2}*{5.0} 
\\
PSNR $\uparrow$  & & & & & & & & \\
\midrule

Mean
& 28.6934
& 31.8259
& 36.0805
& 37.2026
& 37.4394
& 38.7040
& 38.8871
& 38.6065 \\

Std
& 3.3241
& 2.9944
& 2.8315
& 2.9084
& 2.9572
& 3.4891
& 3.2364
& 3.0213 \\

\midrule

 & \multirow{2}*{6.0}
 & \multirow{2}*{8.0}
 & \multirow{2}*{10.0}
 & \multirow{2}*{12.0}
 & \multirow{2}*{14.0}
 & \multirow{2}*{16.0}
 & \multirow{2}*{18.0}
 & \multirow{2}*{20.0} \\
  &  & & & & & & & \\
\midrule
Mean
& 38.5514
& 39.2204
& 39.9796
& \cellcolor{orange}40.1840
& \cellcolor{red}40.2247
& \cellcolor{yellow}39.9558
& 39.1992
& 38.9566 \\

Std
& 2.9264
& 2.9735
& 2.8425
& 2.5163
& 2.3639
& 2.6917
& 2.6648
& 2.9618 \\

\bottomrule
\end{tabular}
\label{tab:image_ablation_paper}
\end{table}

\noindent
\textbf{Findings.}
As shown in Tab.~\ref{tab:image_ablation_paper}, the PSNR steadily increases with \(\beta\) and peaks around \(\beta = 14.0\), after which it saturates or slightly degrades, so we fix \(\beta = 14.0\) for all subsequent 2D image experiments.
With this setting, HOSC clearly improves over PEMLP, Gauss, WIRE, and SIREN, and is competitive with FINER (Tab.~\ref{tab:image_comparison_table}): HOSC matches or exceeds FINER on 9 of the 16 images (see Appendix) while trailing it slightly on average.
These results confirm that reaching a high-\(\beta\) regime is crucial to unlock HOSC’s advantages, and that in this regime HOSC closes most of the gap to the strongest existing periodic activation while delivering substantial gains over standard sine-based INRs.

\begin{table}[bp]
\setlength\tabcolsep{3.5pt}
\footnotesize
\centering
\caption{\textbf{Quantitative comparison on image fitting.} Mean. and Std. values averaged over sixteen test set examples. $\beta = 14.0$.}
\vspace{-1.2em}
\begin{tabular}{ccccccc}
\toprule
 {Example}
&  \multirow{2}*{{PEMLP}} &  \multirow{2}*{{Gauss}} & \multirow{2}*{{WIRE}} & \multirow{2}*{{SIREN}} &   \multirow{2}*{{FINER}} & \multirow{2}*{{HOSC}} \\
PSNR $\uparrow$ & & & & & & \\
\midrule

Temple
& 26.3002
& 31.4108
& 28.0027
& \cellcolor{yellow}33.6837
& \cellcolor{orange}35.8982
& \cellcolor{red}36.6901 \\

Market 
& 30.4807
& 35.8760
& 32.4138
& \cellcolor{yellow}40.2123
& \cellcolor{red}42.6105
& \cellcolor{orange}40.3085 \\

Wallnut
& 24.1596
& 32.2588
& 26.9073
& \cellcolor{yellow}34.4519
& \cellcolor{orange}36.9627
& \cellcolor{red}37.0417 \\

\midrule
Mean
& 29.5525
& 35.6942
& 31.3934
& \cellcolor{yellow}38.5811
& \cellcolor{red} 40.8816
& \cellcolor{orange} 40.2247 \\

Std
& 3.7192
& 2.7251
& 3.0141
& 3.1956
& 3.1379
& 2.3639 \\

\bottomrule
\end{tabular}
\label{tab:image_comparison_table}
\end{table}

\begin{figure*}[htbp]
    \centering
    
    \setlength{\SpaceCaptionToRow}{-0.0em} %
    
    \setlength{\SpaceMetricToNextRow}{-0.0em} %
    
    \setlength{\SpaceImageToMetric}{-1.3em} %

    \vspace{\SpaceCaptionToRow} 

    \def\subfigwidth{0.142\textwidth}

    \begin{subfigure}[t]{\subfigwidth}
        \centering
        {PEMLP} %
        \includegraphics[width=\textwidth]{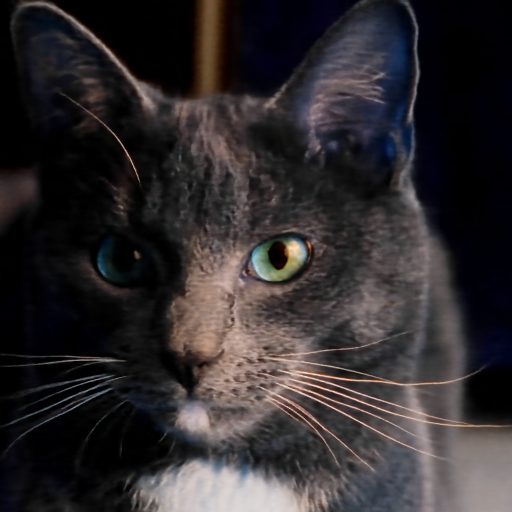}
        \vspace{\SpaceImageToMetric} %
        \caption*{{32.9754dB}} %
    \end{subfigure}%
    \hfill
    \begin{subfigure}[t]{\subfigwidth}
        \centering
        {Gauss} %
        \includegraphics[width=\textwidth]{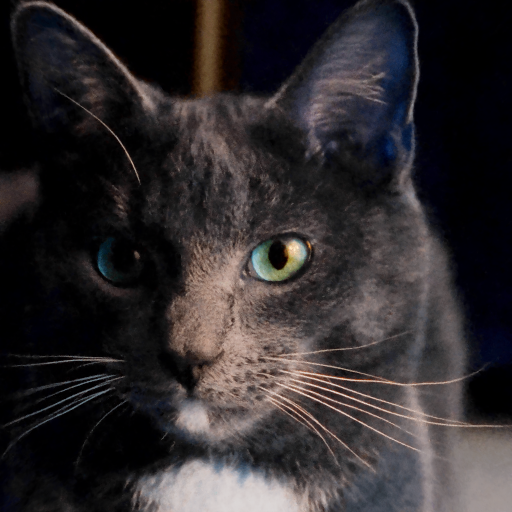}
        \vspace{\SpaceImageToMetric} %
        \caption*{{32.8437dB}} %
    \end{subfigure}%
    \hfill
    \begin{subfigure}[t]{\subfigwidth}
        \centering
        {WIRE} %
        \includegraphics[width=\textwidth]{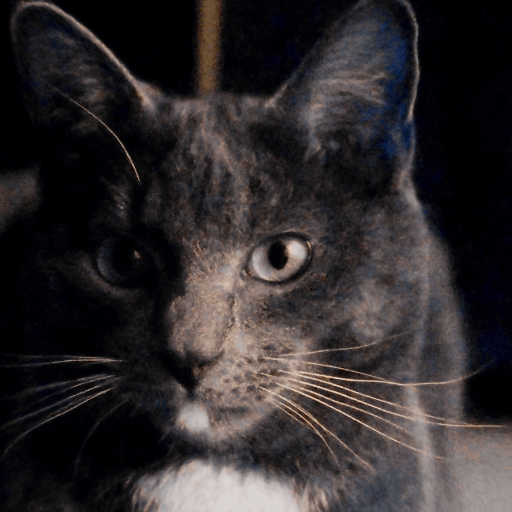}
        \vspace{\SpaceImageToMetric} %
        \caption*{{29.9478dB}} %
    \end{subfigure}%
    \hfill
    \begin{subfigure}[t]{\subfigwidth}
        \centering
        {SIREN} %
        \includegraphics[width=\textwidth]{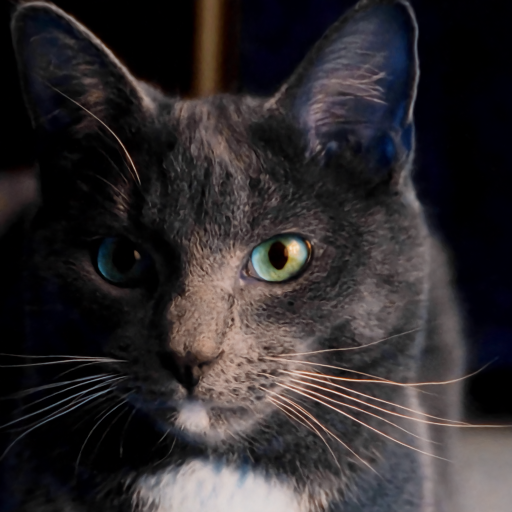}
        \vspace{\SpaceImageToMetric} %
        \caption*{{37.0785dB}} %
    \end{subfigure}%
    \hfill
    \begin{subfigure}[t]{\subfigwidth}
        \centering
        {FINER} %
        \includegraphics[width=\textwidth]{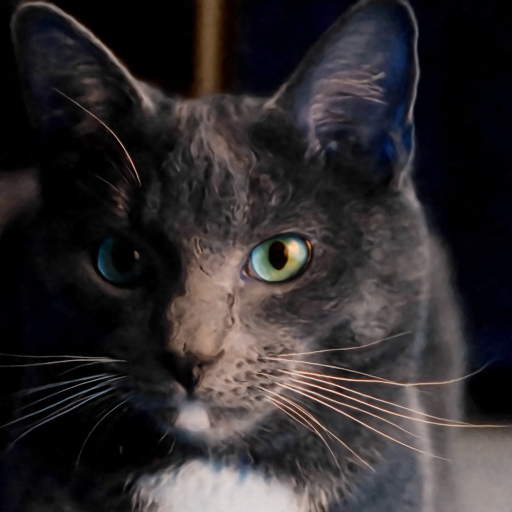}
        \vspace{\SpaceImageToMetric} %
        \caption*{{32.4541dB}} %
    \end{subfigure}%
    \hfill
    \begin{subfigure}[t]{\subfigwidth}
        \centering
        {HOSC} %
        \includegraphics[width=\textwidth]{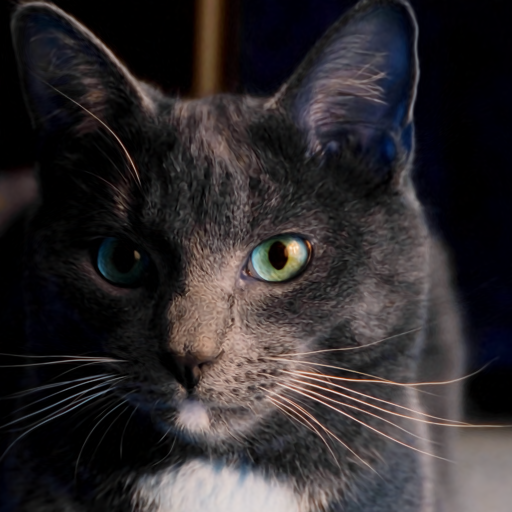}
        \vspace{\SpaceImageToMetric} %
        \caption*{\textbf{40.3450dB}} %
    \end{subfigure}%
    \hfill
    \begin{subfigure}[t]{\subfigwidth}
        \centering
        {GT} %
        \includegraphics[width=\textwidth]{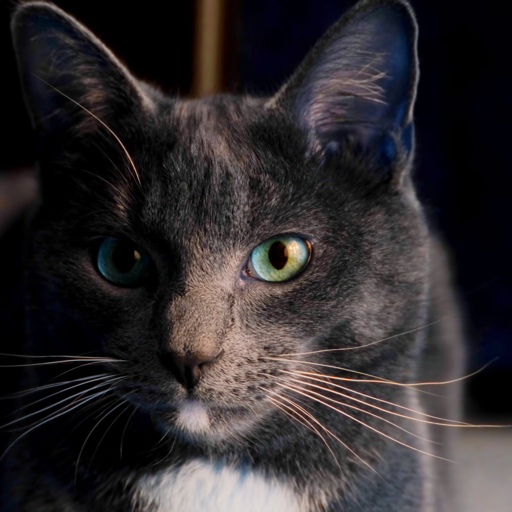} %
        \vspace{\SpaceImageToMetric} %
        \caption*{} %
    \end{subfigure}%
    \hfill

    \vspace{-0.8em}
    \caption{Qualitative comparison on video fitting. A 300 frame video of a cat was reconstructed. Displayed is a single frame from the video. Quantitative results showcased per example below reconstructed frames. HOSC $\beta=0.5$.}
    \label{fig:video_comparison_paper}

\end{figure*}

\begin{figure*}[htbp]
    \centering

    \setlength{\SpaceCaptionToRow}{-0.0em} 
    \setlength{\SpaceMetricToNextRow}{-1.2em} 
    \setlength{\SpaceImageToMetric}{-1.6em} 

    \vspace{\SpaceCaptionToRow}

    \def\subfigwidth{0.142\textwidth}
    
    \newlength{\VerticalLabelWidth}
    \setlength{\VerticalLabelWidth}{0.6cm} 
    
    \newlength{\VerticalLabelHeight}
    \setlength{\VerticalLabelHeight}{2.2cm} 
    \newlength{\VerticalLabelHeightDiff}
    \setlength{\VerticalLabelHeightDiff}{1.8cm} 

    \begin{tabular}{@{}c@{ }c@{}}
        \mbox{\hspace{\VerticalLabelWidth}} & %
        \resizebox{0.95\textwidth}{!}{%
        \begin{minipage}[t]{\textwidth}
            \centering
            \begin{minipage}{\subfigwidth}
                \centering{PEMLP}
            \end{minipage}%
            \hfill
            \begin{minipage}{\subfigwidth}
                \centering{Gauss}
            \end{minipage}%
            \hfill
            \begin{minipage}{\subfigwidth}
                \centering{WIRE}
            \end{minipage}%
            \hfill
            \begin{minipage}{\subfigwidth}
                \centering{SIREN}
            \end{minipage}%
            \hfill
            \begin{minipage}{\subfigwidth}
                \centering{FINER}
            \end{minipage}%
            \hfill
            \begin{minipage}{\subfigwidth}
                \centering{HOSC}
            \end{minipage}%
            \hfill
            \begin{minipage}{\subfigwidth}
                \centering{GT}
            \end{minipage}%
        \end{minipage}%
        }%
        \\ 
        \noalign{\vspace{\SpaceMetricToNextRow}}

        \rotatebox{90}{\parbox{\VerticalLabelHeight}{\centering{Bach\\Audio}}} &
        \resizebox{0.95\textwidth}{!}{%
        \begin{minipage}[t]{\textwidth}
            \centering
            \begin{subfigure}[t]{\subfigwidth}
                \centering
                \includegraphics[width=\textwidth]{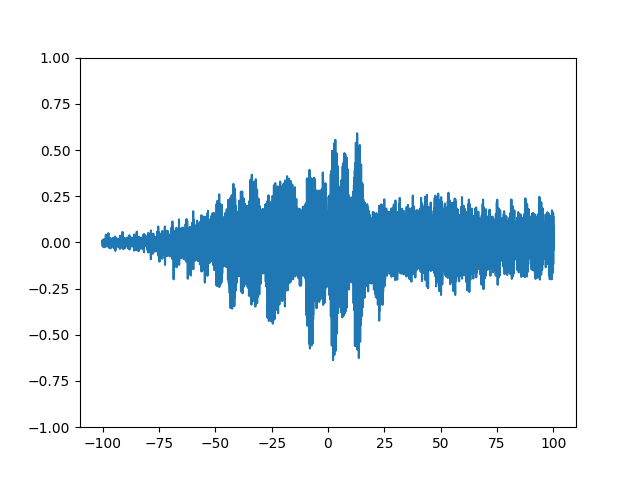}
                \vspace{\SpaceImageToMetric}
                \caption*{}
            \end{subfigure}%
            \hfill
            \begin{subfigure}[t]{\subfigwidth}
                \centering
                \includegraphics[width=\textwidth]{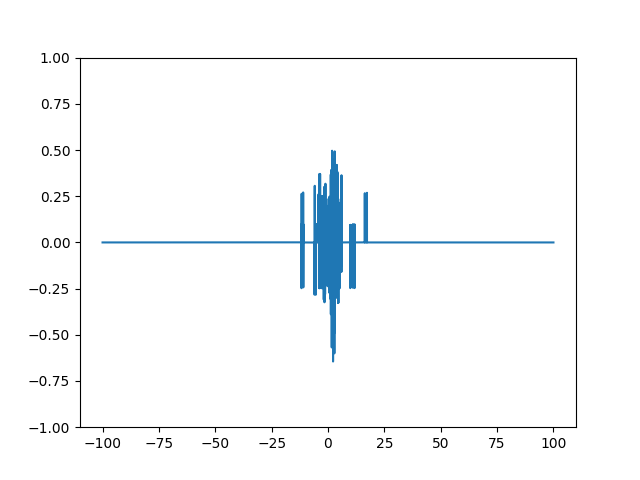}
                \vspace{\SpaceImageToMetric}
                \caption*{}
            \end{subfigure}%
            \hfill
            \begin{subfigure}[t]{\subfigwidth}
                \centering
                \includegraphics[width=\textwidth]{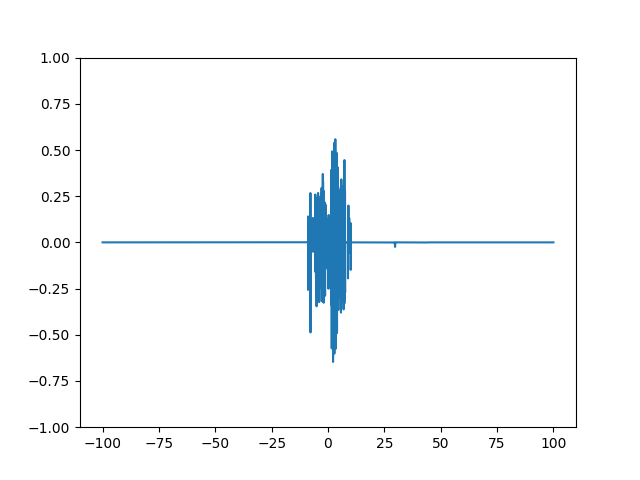}
                \vspace{\SpaceImageToMetric}
                \caption*{}
            \end{subfigure}%
            \hfill
            \begin{subfigure}[t]{\subfigwidth}
                \centering
                \includegraphics[width=\textwidth]{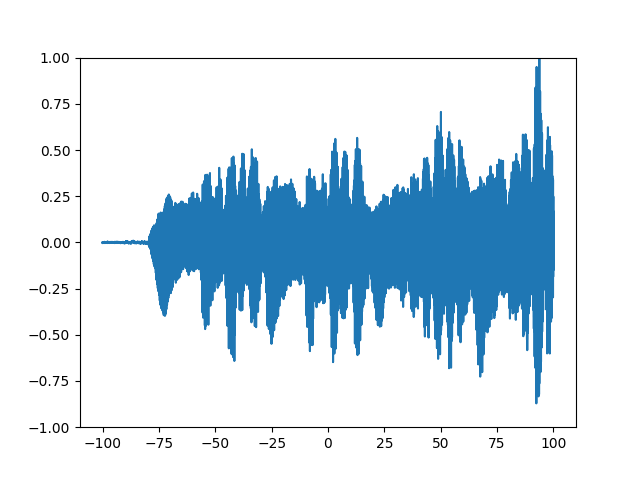}
                \vspace{\SpaceImageToMetric}
                \caption*{}
            \end{subfigure}%
            \hfill
            \begin{subfigure}[t]{\subfigwidth}
                \centering
                \includegraphics[width=\textwidth]{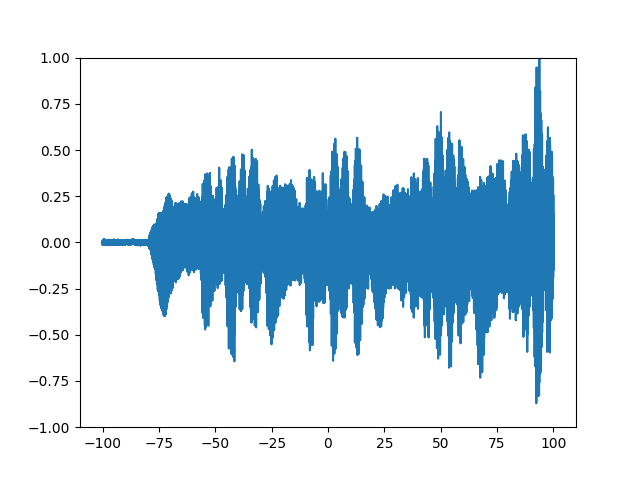}
                \vspace{\SpaceImageToMetric}
                \caption*{}
            \end{subfigure}%
            \hfill
            \begin{subfigure}[t]{\subfigwidth}
                \centering
                \includegraphics[width=\textwidth]{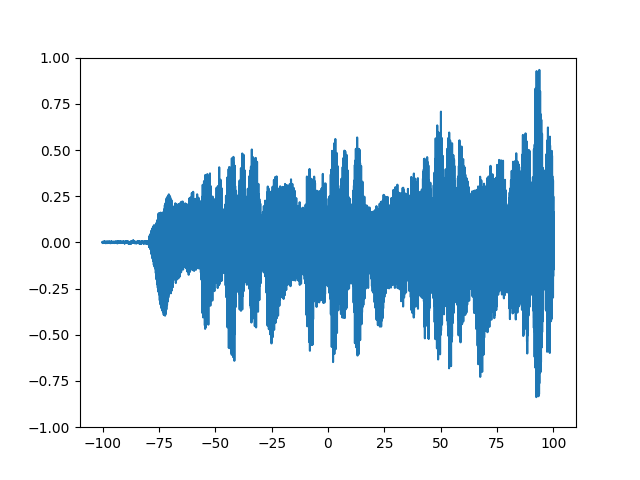}
                \vspace{\SpaceImageToMetric}
                \caption*{}
            \end{subfigure}%
            \hfill
            \begin{subfigure}[t]{\subfigwidth}
                \centering
                \includegraphics[width=\textwidth]{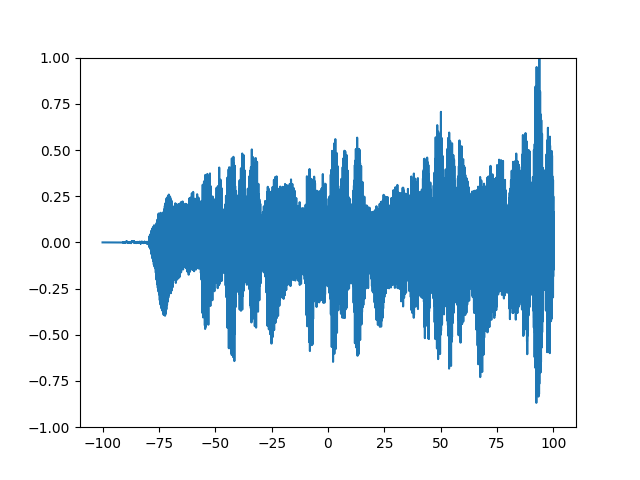}
                \vspace{\SpaceImageToMetric}
                \caption*{}
            \end{subfigure}%
            \hfill
        \end{minipage}%
        }%
        \\ 
        \noalign{\vspace{\SpaceMetricToNextRow}}

        \rotatebox{90}{\parbox{\VerticalLabelHeightDiff}{\centering{Bach\\Diff}}}
        &
        \resizebox{0.95\textwidth}{!}{%
        \begin{minipage}[t]{\textwidth}
            \begin{subfigure}[t]{\subfigwidth}
                \centering
                \includegraphics[width=\textwidth]{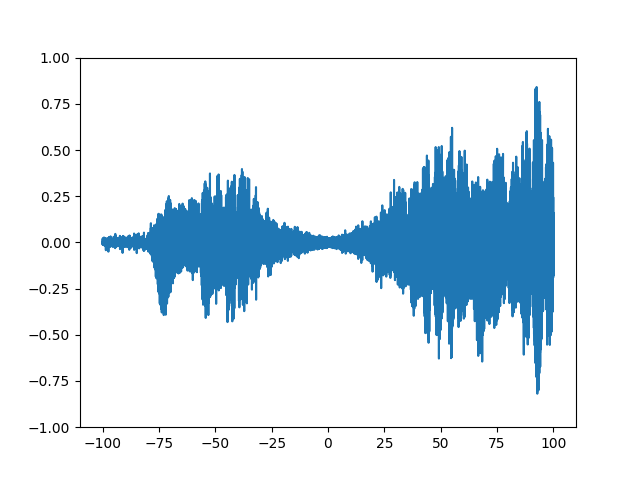}
                \vspace{\SpaceImageToMetric}
                \caption*{{1.3899e-02}}
            \end{subfigure}%
            \hfill
            \begin{subfigure}[t]{\subfigwidth}
                \centering
                \includegraphics[width=\textwidth]{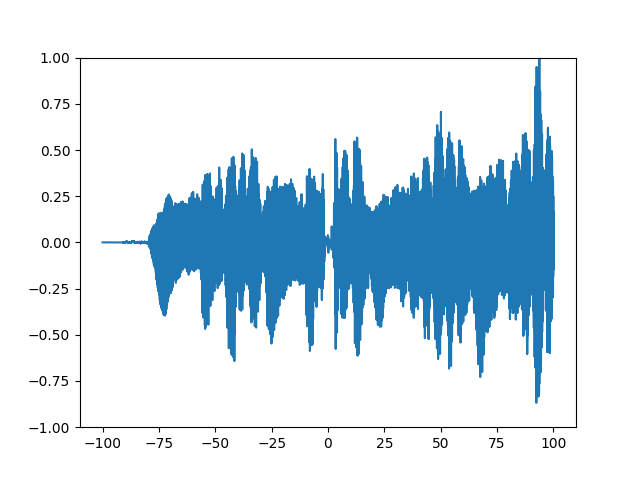}
                \vspace{\SpaceImageToMetric}
                \caption*{{2.3240e-02}}
            \end{subfigure}%
            \hfill
            \begin{subfigure}[t]{\subfigwidth}
                \centering
                \includegraphics[width=\textwidth]{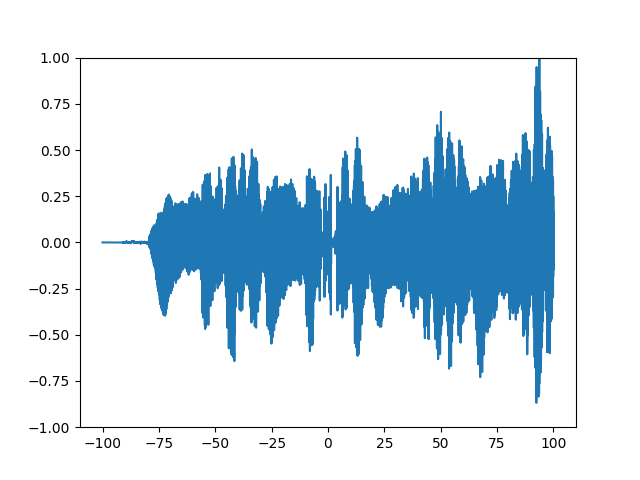}
                \vspace{\SpaceImageToMetric}
                \caption*{{2.2847e-02}}
            \end{subfigure}%
            \hfill
            \begin{subfigure}[t]{\subfigwidth}
                \centering
                \includegraphics[width=\textwidth]{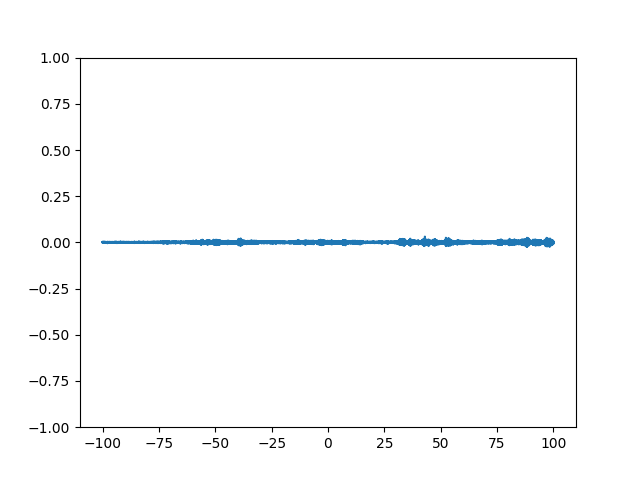}
                \vspace{\SpaceImageToMetric}
                \caption*{{5.6026e-06}}
            \end{subfigure}%
            \hfill
            \begin{subfigure}[t]{\subfigwidth}
                \centering
                \includegraphics[width=\textwidth]{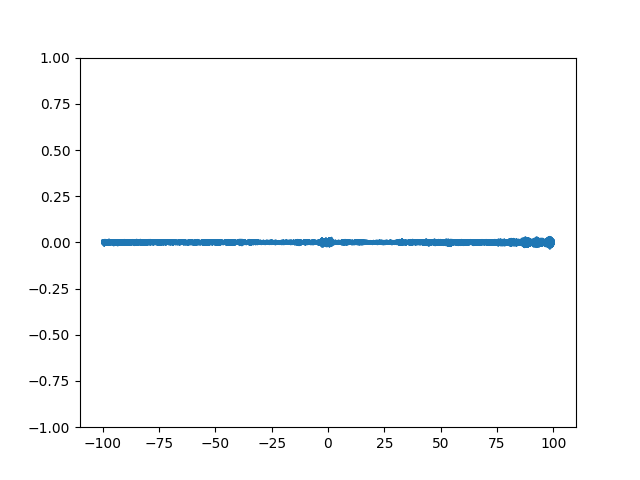}
                \vspace{\SpaceImageToMetric}
                \caption*{{1.2803e-05}}
            \end{subfigure}%
            \hfill
            \begin{subfigure}[t]{\subfigwidth}
                \centering
                \includegraphics[width=\textwidth]{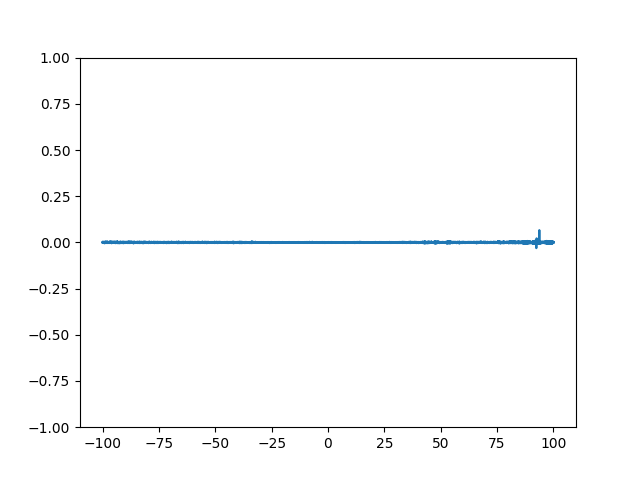}
                \vspace{\SpaceImageToMetric}
                \caption*{\textbf{1.3194e-06}}
            \end{subfigure}%
            \hfill
            \begin{subfigure}[t]{\subfigwidth}
                \centering
                \vspace{\SpaceImageToMetric}
                \caption*{}
            \end{subfigure}%
            \hfill
        \end{minipage}%
        }%
        \\

    \end{tabular}
    \vspace{-0.8em}
    \caption{Qualitative comparison on audio fitting. We visualize the reconstructed signal (Audio) and the difference of predicted and GT signal (Diff). We quantify the difference with MSE. HOSC $\beta=5.0$.}
    \label{fig:audio_comparison_final_two_rows}

\end{figure*}

\subsection{Audio Fitting}
We evaluate the effectiveness of HOSC on the task of fitting an MLP to audio signals. The MLP learns an implicit 1D function, $f\colon\mathbb{R}^{1} \rightarrow \mathbb{R}^{1}$ of signal amplitude with respect to time. The loss function is MSE. We evaluate the reconstruction quality on two examples, the first 7 seconds of Bach’s Cello Suite No. 1: Prelude (Bach) and a 12 second clip
of a male actor counting 0-9 (Counting). We present Bach results in the paper and Counting in the Appendix. %

\noindent
\textbf{Findings.}
The \(\beta\)-ablation in Tab.~\ref{tab:audio_ablation} shows that \(\beta\) in the range \(4\)–\(5\) yields the best reconstruction quality across both signals; we fix \(\beta = 5.0\) as a single setting for all audio experiments. With this choice, HOSC substantially improves over all baselines (Tab.~\ref{tab:audio_comparison}): compared to the next best activation (SIREN), HOSC reduces MSE by more than \(4\times\) on Bach and nearly an order of magnitude on Counting. This makes the audio setting the clearest example where entering the high-\(\beta\) regime unlocks a level of fidelity that sinusoidal activations alone cannot reach.

\begin{table}[bp]
\centering
\scriptsize
\setlength\tabcolsep{3.05pt}
\caption{\textbf{Audio HOSC $\beta$-ablation}. Showcased for Bach and Counting examples. Full ablation in the Appendix}
\label{tab:audio_ablation}

\vspace{-1.2em}
\begin{tabular}{ccccccc}
\toprule
& \multicolumn{6}{c}{\textbf{HOSC } $\boldsymbol{\beta}$} \\
\cmidrule(lr){2-7}
Example & 
\multirow{2}*{1.0} &
\multirow{2}*{2.0} &
\multirow{2}*{4.0} &
\multirow{2}*{5.0} &
\multirow{2}*{6.0} &
\multirow{2}*{8.0} \\
MSE $\downarrow$ &  &  &  &  &  &  \\

\midrule
Bach &  
9.86e-06 & \cellcolor{yellow}2.029e-06 & \cellcolor{red}7.589e-07 & \cellcolor{orange}1.319e-06 & 2.384e-06 & 2.138e-05 \\

Counting & 
4.328e-04 & 2.550e-04 & \cellcolor{orange}6.851e-05 & \cellcolor{red}3.613e-05 & \cellcolor{yellow}9.355e-05 & 1.707e-04

 \\ 
\bottomrule
\end{tabular}
\end{table}

\begin{table}[bp]
\setlength\tabcolsep{1.6pt}
\scriptsize
\centering
\caption{\textbf{Quantitative comparisons on audio fitting.} We report lowest achieved MSE loss on each audio sample. $\beta=5.0$}
\label{tab:audio_comparison}
\vspace{-1.2em}
\begin{tabular}{ccccccc}
\toprule
 {Example}
&  \multirow{2}*{{PEMLP}} &  \multirow{2}*{{Gauss}} & \multirow{2}*{{WIRE}} & \multirow{2}*{{SIREN}} &   \multirow{2}*{{FINER}} & \multirow{2}*{{HOSC}} \\
MSE $\downarrow$ & & & & & & \\
\midrule

Bach 
& 1.3899e-02
& 2.3240e-02
& 2.2847e-02
& \cellcolor{orange}5.6026e-06
& \cellcolor{yellow}1.2803e-05
& \cellcolor{red}1.3194e-06 \\

Counting
& 4.1440e-03
& 7.4262e-03
& 7.4241e-03
& \cellcolor{orange}3.9236e-04
& \cellcolor{yellow}5.9250e-03
& \cellcolor{red}3.6126e-05 \\
\bottomrule
\end{tabular}
\end{table}

\subsection{Video Fitting}
We evaluate HOSC’s effectiveness on video reconstruction, where an MLP parameterizes a function \(f: \mathbb{R}^{3} \rightarrow \mathbb{R}^{3}\) mapping \((x,y,t)\) to RGB values. The loss is mean squared error (MSE). We use a 300-frame cat video at \(512 \times 512\) resolution and report per-frame and average PSNR/MSE. All methods share the same network architecture, optimizer, training schedule, and initialization; only the activation (and \(\beta\) for HOSC) differs. We first ablate \(\beta\) and then compare HOSC to PEMLP, Gauss, WIRE, SIREN, and FINER using the best \(\beta\).

\paragraph{Findings.}
The \(\beta\)-ablation in Tab.~\ref{tab:video_ablation} shows a clear optimum at \(\beta = 0.5\); smaller or larger values reduce reconstruction quality. With this choice, HOSC achieves the best overall performance (Tab.~\ref{tab:video_comparison}): it improves PSNR by about \(3\) dB and halves the MSE relative to SIREN, while non-periodic baselines lag further behind. This aligns with our analysis that (cf. Sec.~\ref{sec:asymp_behavior}), for higher-dimensional coordinate inputs such as \((x,y,t)\), a relatively small \(\beta\) keeps HOSC in a smoothly saturated, sine-like regime that yields stable and accurate reconstructions.

\begin{table}[bp]
\centering
\scriptsize
\setlength{\tabcolsep}{4.6pt}
\caption{\textbf{Video HOSC $\beta$-ablation}. Showcased for the Cat video.}
\label{tab:video_ablation}
\vspace{-1.2em}
\begin{tabular}{cccccc}
\toprule
& \multicolumn{5}{c}{\textbf{HOSC }$\boldsymbol{\beta}$} \\
\cmidrule(lr){2-6}
Metric & 0.1 & 0.5 & 1.0 & 2.0 & 5.0 
\\

\midrule

MSE $\downarrow$
& \cellcolor{yellow}7.7876e-04
& \cellcolor{red}5.5673e-04
& \cellcolor{orange}6.7687e-04
& 1.5408e-02
& 1.6507e-01 \\

PSNR $\uparrow$
& \cellcolor{yellow}37.3071
& \cellcolor{red}38.71367
& \cellcolor{orange}37.821784
& 24.2534
& 13.9256 \\

\bottomrule

\end{tabular}
\end{table}

\begin{table}[bp]
\setlength\tabcolsep{1.8pt}
\scriptsize
\centering
\caption{Quantitative comparisons on video fitting. $\beta=0.5$}
\label{tab:video_comparison}
\vspace{-1.2em}
\begin{tabular}{ccccccc}
\toprule
 {Metric}
&  {PEMLP} &  {Gauss} & {WIRE} & {SIREN} & {FINER} & {HOSC} \\
\midrule

MSE $\downarrow$
& 2.3615e-03
& 1.8921e-03
& 3.8377e-03
& \cellcolor{orange}1.0487e-03
& \cellcolor{yellow}2.4904e-03
& \cellcolor{red}5.5673e-04 \\

PSNR $\uparrow$
& 32.3877
& 33.3167
& 30.2326
& \cellcolor{orange}35.9106
& \cellcolor{yellow}32.1581
& \cellcolor{red}38.7137 \\

\bottomrule
\end{tabular}
\end{table}

\begin{figure*}[ht]
    \centering
    
    \setlength{\SpaceCaptionToRow}{-0.0em} %
    
    \setlength{\SpaceMetricToNextRow}{-1em} %
    
    \setlength{\SpaceImageToMetric}{-2.4em} %

    \vspace{\SpaceCaptionToRow} 

    \def\subfigwidth{0.142\textwidth}

    \hfill
    \begin{subfigure}[t]{\subfigwidth}
        \centering
        {PEMLP} %
        \includegraphics[width=\textwidth]{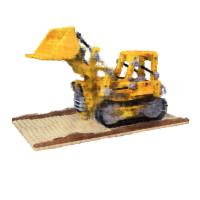}
        \vspace{\SpaceImageToMetric} %
        \caption*{{0.893}} %
    \end{subfigure}%
    \hfill
    \begin{subfigure}[t]{\subfigwidth}
        \centering
        {Gauss} %
        \includegraphics[width=\textwidth]{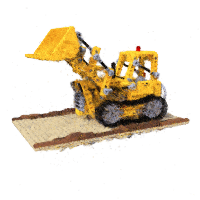}
        \vspace{\SpaceImageToMetric} %
        \caption*{{0.892}} %
    \end{subfigure}%
    \hfill
    \begin{subfigure}[t]{\subfigwidth}
        \centering
        {WIRE} %
        \includegraphics[width=\textwidth]{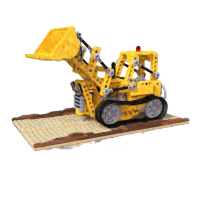}
        \vspace{\SpaceImageToMetric} %
        \caption*{{0.949}} %
    \end{subfigure}%
    \hfill
    \begin{subfigure}[t]{\subfigwidth}
        \centering
        {SIREN} %
        \includegraphics[width=\textwidth]{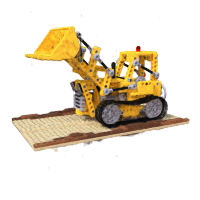}
        \vspace{\SpaceImageToMetric} %
        \caption*{{0.958}} %
    \end{subfigure}%
    \hfill
    \begin{subfigure}[t]{\subfigwidth}
        \centering
        {FINER} %
        \includegraphics[width=\textwidth]{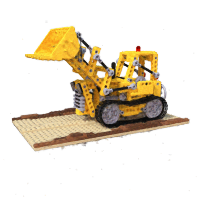}
        \vspace{\SpaceImageToMetric} %
        \caption*{{0.959}} %
    \end{subfigure}%
    \hfill
    \begin{subfigure}[t]{\subfigwidth}
        \centering
        {HOSC} %
        \includegraphics[width=\textwidth]{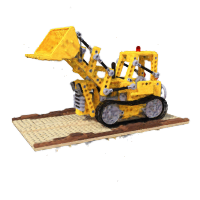}
        \vspace{\SpaceImageToMetric} %
        \caption*{\textbf{0.960}} %
    \end{subfigure}%
    \hfill
    \begin{subfigure}[t]{\subfigwidth}
        \centering
        {GT} %
        \includegraphics[width=\textwidth]{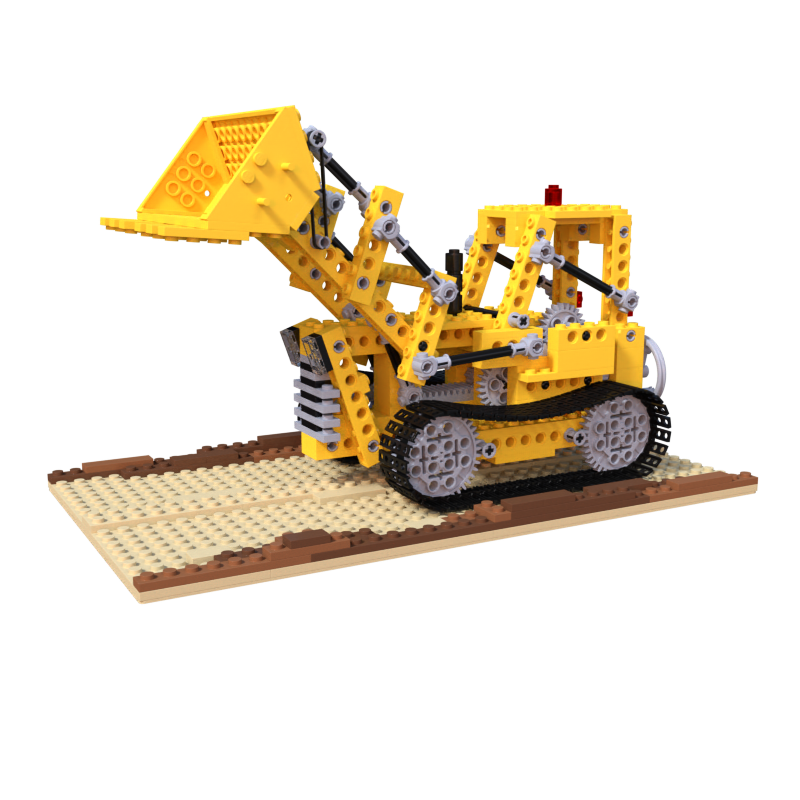} %
        \vspace{\SpaceImageToMetric} %
        \caption*{} %
    \end{subfigure}%
    \hfill
    \vspace{-0.8em}
    \caption{Qualitative comparison on image fitting.
    SSIM metric showcased under each method for the Lego truck example from the NeRF Synthetic dataset\cite{mildenhall2020nerf}.
    HOSC $\beta=1.0$. }
    \label{fig:nerf_comparison_paper}
    
\end{figure*}

\begin{figure*}[tp]
    \centering
    
    \setlength{\SpaceCaptionToRow}{-0.0em} %
    
    \setlength{\SpaceMetricToNextRow}{-0.2em} %
    
    \setlength{\SpaceImageToMetric}{-1.3em} %

    \vspace{\SpaceCaptionToRow} 

    \def\subfigwidth{0.133\textwidth} 
    \hfill
    \begin{subfigure}[t]{\subfigwidth}
        \centering
        {Gauss} %
        \includegraphics[trim={3cm 0 3cm 0},clip,width=\textwidth]{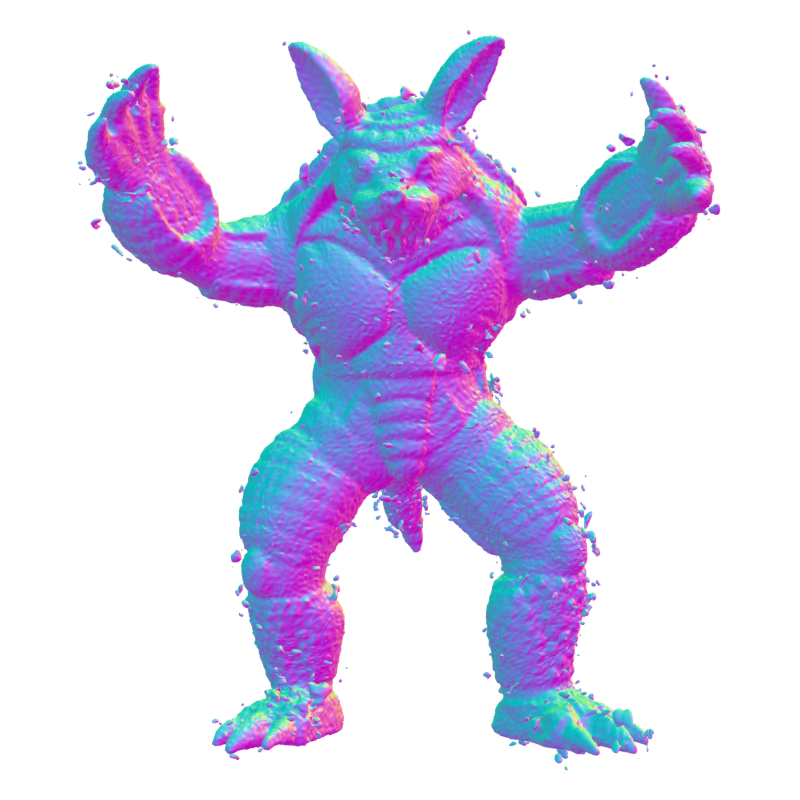}
        \vspace{\SpaceImageToMetric} %
        \caption*{{2.2203e-05}} %
    \end{subfigure}%
    \hfill
    \begin{subfigure}[t]{\subfigwidth}
        \centering
        {WIRE} %
        \includegraphics[trim={3cm 0 3cm 0},clip,width=\textwidth]{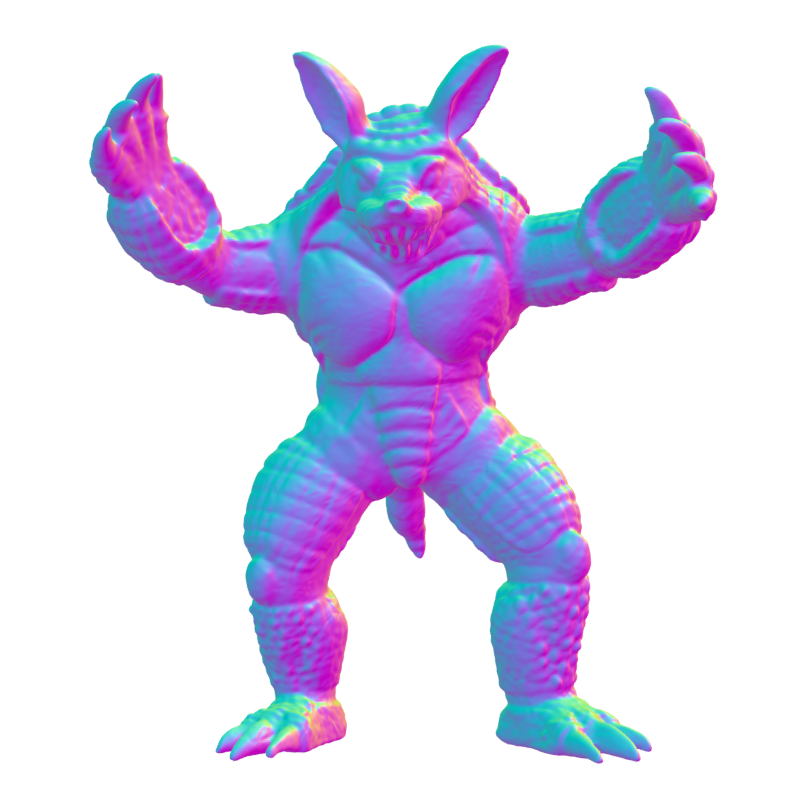}
        \vspace{\SpaceImageToMetric} %
        \caption*{{4.4079e-06}} %
    \end{subfigure}%
    \hfill
    \begin{subfigure}[t]{\subfigwidth}
        \centering
        {SIREN} %
        \includegraphics[trim={3cm 0 3cm 0},clip,width=\textwidth]{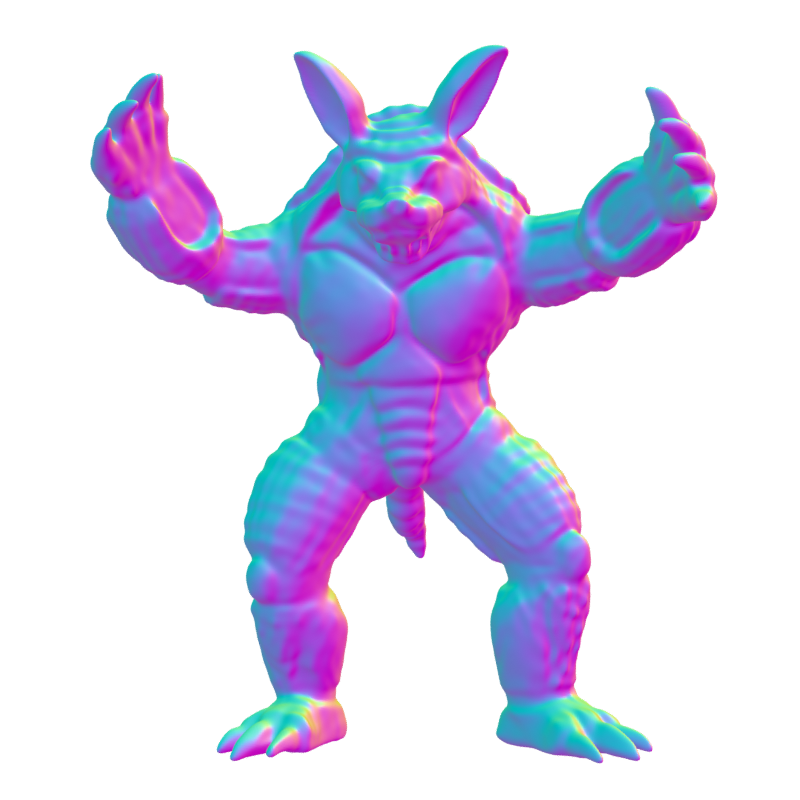}
        \vspace{\SpaceImageToMetric} %
        \caption*{{4.9024e-06}} %
    \end{subfigure}%
    \hfill
    \begin{subfigure}[t]{\subfigwidth}
        \centering
        {FINER} %
        \includegraphics[trim={3cm 0 3cm 0},clip,width=\textwidth]{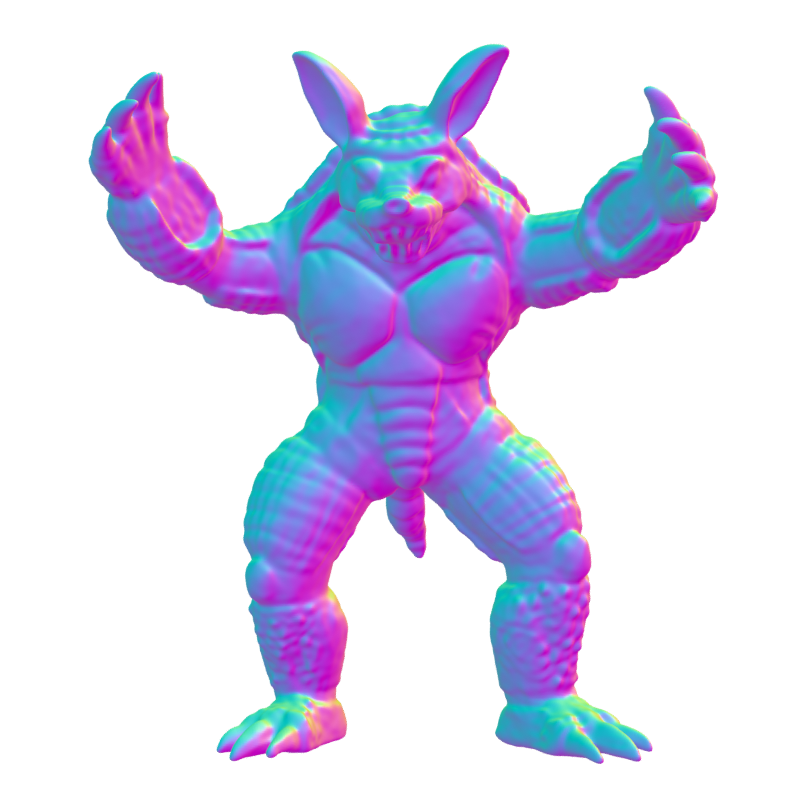} %
        \vspace{\SpaceImageToMetric} %
        \caption*{\textbf{4.3120e-06}} %
    \end{subfigure}%
    \hfill
    \begin{subfigure}[t]{\subfigwidth}
        \centering
        {HOSC} %
        \includegraphics[trim={3cm 0 3cm 0},clip,width=\textwidth]{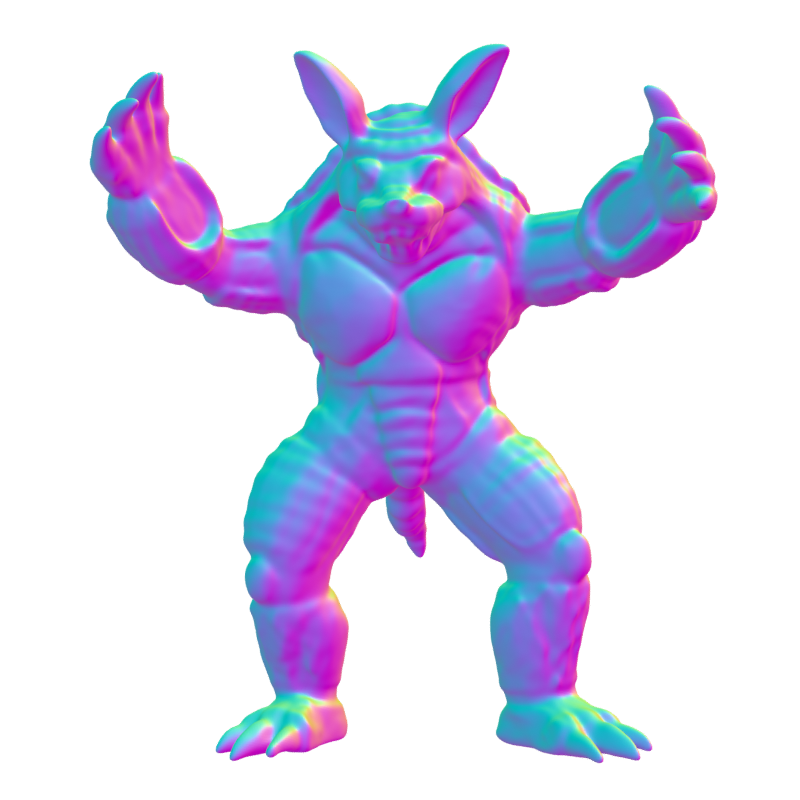} %
        \vspace{\SpaceImageToMetric} %
        \caption*{{4.4470-06}} %
    \end{subfigure}%
    \hfill
        \begin{subfigure}[t]{\subfigwidth}
        \centering
        {GT} %
        \includegraphics[trim={3cm 0 3cm 0},clip,width=\textwidth]{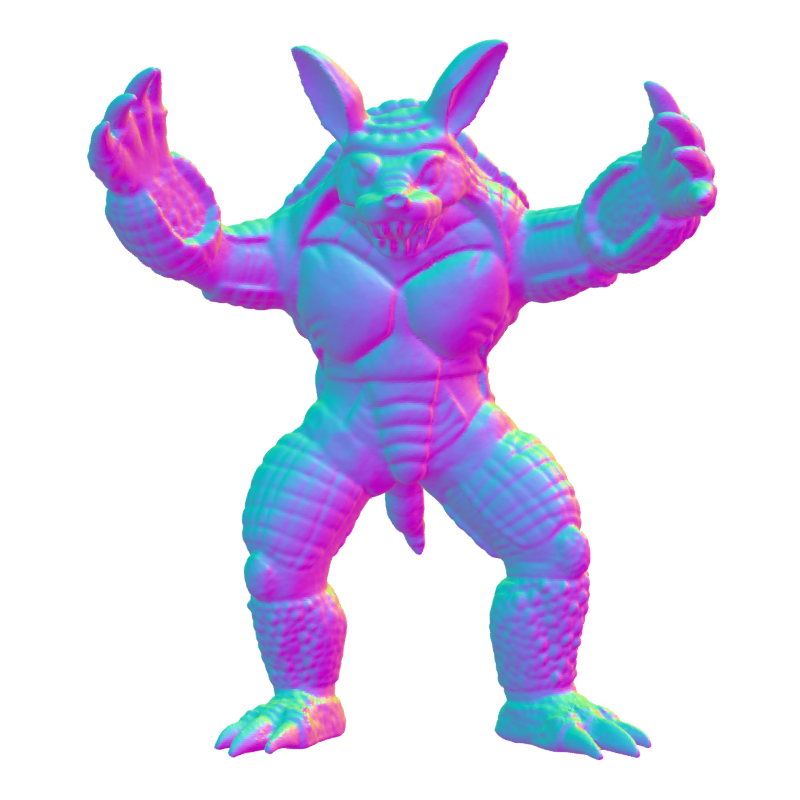} %
        \vspace{\SpaceImageToMetric} %
        \caption*{} %
    \end{subfigure}%
    \hfill
    \vspace{-0.8em}
    \caption{Qualitative comparison on signed distance field (Neural SDF) fitting.
    Chamfer Distance showcased per example below the reconstructed shape. $\beta=0.8$. We provide full results, along with an ablation study and experimental details in the Appendix.}
    \label{fig:sdf_comparison_paper}
    
\end{figure*}

\subsection{Neural Radiance Fields (NeRF)}

We evaluate HOSC and other activations based methods on the Blender benchmark (8 scenes) \cite{mildenhall2020nerf}.
All methods share the same architecture, optimizer, training budget, and SIREN-style initialization; no positional encodings are used.
We ablate the HOSC $\beta$ parameter 
and compare under identical budgets on  PSNR, SSIM and LPIPS metrics.

\noindent
\textbf{Findings.}
The \(\beta\)-ablation in Tab.~\ref{tab:nerf_ablation} shows that NeRF performance peaks for \(\beta \in [0.5,1.0]\); larger values quickly degrade PSNR/SSIM and LPIPS, with \(\beta \ge 3\) leading to severe collapse across all scenes (Tab.~\ref{tab:nerf_ablation}). With \(\beta\) chosen in this low range, HOSC matches SIREN and FINER on the Blender benchmark (Tab.~\ref{tab:res_nerf}), with per-scene PSNR differences on the order of \(0.1\)–\(0.2\)\,dB (FINER 29.84\,dB, HOSC 29.74\,dB, SIREN 29.71\,dB). In NeRF, the main benefit of HOSC is thus a safe, explicit gradient-scale knob that preserves the frequency support of sine-based INRs, rather than a new accuracy regime as in our audio experiments.

\begin{table}[bp]
\centering
\scriptsize
\setlength{\tabcolsep}{3.1pt}
\renewcommand{\arraystretch}{1.2}
\caption{\textbf{NeRF HOSC $\beta$-ablation}. Averaged over eight sample from NeRF Synthetic \cite{mildenhall2020nerf}}
\vspace{-1.2em}
\begin{tabular}{ccccccccc}
\toprule
 & & \multicolumn{7}{c}{\textbf{HOSC } $\boldsymbol{\beta}$} \\
\cmidrule(lr){3-9}
&
Stat. & 
0.1 & 0.3 &
0.5 & 0.8 &
1.0 & 3.0 &
5.0 \\

\midrule

\multirow{2}*{\rotatebox{90}{PSNR}} 
& Mean.
& 26.4848
& 28.7892
& \cellcolor{yellow}29.3195
& \cellcolor{orange}29.6461
& \cellcolor{red}29.7321
& 26.7590
& 21.6911 \\

& Std.
& 4.5546
& 4.4083
& 4.4643
& 4.6295
& 4.6157
& 4.0048
& 3.1601 \\
\cmidrule(lr){2-9}

\multirow{2}*{\rotatebox{90}{SSIM}} &
Mean.
& 0.9002
& 0.9300
& \cellcolor{yellow}0.9346
& \cellcolor{orange}0.9372
& \cellcolor{red}0.9388
& 0.9033
& 0.7999 \\

& Std.
& 0.0765
& 0.0652
& 0.0629
& 0.0640
& 0.0622
& 0.0579
& 0.0820 \\

\cmidrule(lr){2-9}

\multirow{2}*{\rotatebox{90}{LPIPS}} &
Mean.
& 0.1157
& 0.0455
& \cellcolor{yellow}0.0393
& \cellcolor{orange}0.0358
& \cellcolor{red}0.0353
& 0.0757
& 0.2355 \\ 

& Std.
& 0.0835
& 0.0374
& 0.0318
& 0.0282
& 0.0277
& 0.0427
& 0.0768 \\

\bottomrule
\end{tabular}
\label{tab:nerf_ablation}
\end{table}

\begin{table}[bp]
\setlength\tabcolsep{3.0pt}
\scriptsize
\centering
\caption{\textbf{Quantitative comparisons on novel view synthesis}. HOSC $\beta=1.0$ }
\vspace{-1.2em}
\begin{tabular}{cccccccccc}
\toprule
 &{Methods}
& {Chair} & {Drums} & {Ficus} & {Hotdog} 
& {Lego} & {Materials} & {Mic} & {Ship} \\
\midrule
\multirow{5}*{\rotatebox{90}{PSNR $\uparrow$}} 
& {PEMLP} & 29.33& 20.11& 23.33& 28.32& 24.56& 23.80& 22.59& 19.84 \\
& {Gauss} & 29.33 & 22.51 & 25.45 & 29.59 & 25.98 & 24.87 & 31.51 & 20.36 \\
& {WIRE}  & 33.86 & 25.08 & 27.85 & 34.18 & 29.41 & 27.85 & 34.95 & 20.61 \\ 
& {SIREN}  & \cellcolor{orange}34.39 & \cellcolor{yellow}25.75 & \cellcolor{orange}28.45 & \cellcolor{red}33.96 & \cellcolor{yellow}30.26 & \cellcolor{orange}27.81 & \cellcolor{orange}35.08 & \cellcolor{orange}21.99 \\
& {FINER}  & \cellcolor{red}34.77 & \cellcolor{red}25.91 & \cellcolor{red}28.86 & \cellcolor{yellow}33.58 & \cellcolor{red}30.76 & \cellcolor{yellow}27.74 & \cellcolor{yellow}35.05 & \cellcolor{red}22.04 \\
& {HOSC}   & \cellcolor{yellow}34.28 & \cellcolor{orange}25.83 & \cellcolor{yellow}28.36 & \cellcolor{orange}33.77 & \cellcolor{orange}30.49 & \cellcolor{red}27.86 & \cellcolor{red}35.28 & \cellcolor{yellow}21.98 \\

\midrule
\multirow{5}*{\rotatebox{90}{SSIM $\uparrow$}} 
&{PEMLP} & 0.950 & 0.823 & 0.907 & 0.925 & 0.893 & 0.907 & 0.931 & 0.712 \\           
& {Gauss}  & 0.940 & 0.851 & 0.932 & 0.919 & 0.892 & 0.902 & 0.970 & 0.696 \\         
& WIRE   & 0.976 & 0.915 & 0.958 & 0.971 & 0.949 & 0.945 & 0.982 & 0.747 \\
& {SIREN}  & \cellcolor{red}0.978 & \cellcolor{orange}0.923 & \cellcolor{yellow}0.960 & \cellcolor{red}0.969 & \cellcolor{yellow}0.958 & \cellcolor{red}0.946 & \cellcolor{orange}0.982 & 0.\cellcolor{orange}783 \\           
& {FINER}  & \cellcolor{orange}0.977 & \cellcolor{orange}0.923 & \cellcolor{red}0.962 & \cellcolor{yellow}0.967 & \cellcolor{orange}0.959 & \cellcolor{orange}0.943 & \cellcolor{orange}0.982 & 0.\cellcolor{yellow}781 \\ 
& {HOSC}  & \cellcolor{orange}0.977 & \cellcolor{red}0.925 & \cellcolor{orange}0.961 & \cellcolor{orange}0.968 & \cellcolor{red}0.960 & \cellcolor{red}0.946 & \cellcolor{red}0.983 & 0.\cellcolor{red}792 \\

\midrule
\multirow{5}*{\rotatebox{90}{LPIPS $\downarrow$}} 
& {PEMLP} & 0.039 & 0.180 & 0.078 & 0.054 & 0.085 & 0.069 & 0.121 & 0.236 \\           
&{Gauss}  & 0.042 & 0.109 & 0.057 & 0.082 & 0.061 & 0.086 & 0.024 & 0.224 \\
&WIRE   & 0.015 & 0.053 & 0.033 & 0.018 & 0.034 & 0.032 & 0.015 & 0.187 \\    
&{SIREN}  & \cellcolor{orange}0.014 & 0\cellcolor{yellow}.046 & \cellcolor{yellow}0.029 & \cellcolor{red}0.023 & \cellcolor{yellow}0.024 & \cellcolor{red}0.030 & \cellcolor{orange}0.0132 & \cellcolor{orange}0.106 \\
&{FINER}  & \cellcolor{red}0.012 & \cellcolor{red}0.043 & \cellcolor{red}0.027 & \cellcolor{yellow}0.028 & \cellcolor{red}0.021 & \cellcolor{orange}0.031 & \cellcolor{red}0.011 & \cellcolor{yellow}0.108\\
&{HOSC}  & \cellcolor{orange}0.014 & \cellcolor{orange}0.045 & \cellcolor{red}0.030 & \cellcolor{orange}0.027 & \cellcolor{orange}0.023 & \cellcolor{orange}0.031 & \cellcolor{yellow}0.014 & \cellcolor{red}0.099 \\
          
\bottomrule
\end{tabular}
\label{tab:res_nerf}
\end{table}

\subsection{Signed Distance Fields}

We next test HOSC on 3D signed distance field (SDF) reconstruction, where an MLP parameterizes a function \(f\colon\mathbb{R}^3 \rightarrow \mathbb{R}\) whose zero-level set defines a surface. Following prior work~\cite{lindell2022bacon,zhuang2023antialias}, we train all methods to minimize a point-to-surface loss and report Chamfer distance on held-out surface samples.
We show a representative qualitative comparison on the \texttt{Arm\-a\-dil\-lo} shape in Fig.~\ref{fig:sdf_comparison_paper}, together with per-method Chamfer distances. Additional shapes and a small \(\beta\)-sweep for HOSC are reported in the Appendix.

\paragraph{Findings.}
On \texttt{Arm\-a\-dil\-lo}, HOSC achieves a chamfer distance that is on par with periodic baselines and clearly improves over non-periodic activations (Fig.~\ref{fig:sdf_comparison_paper}): Gaussian is an order of magnitude worse, while WIRE, SIREN, FINER, and HOSC all lie in a narrow band around. %

\section{Conclusions}

We introduced HOSC, a periodic activation that decouples gradient scale from the carrier frequency and adds an explicit gradient–gating mechanism. Analytically, we derived tight activation-level Lipschitz bounds, showed that the high-gradient region shrinks as $O(1/\beta)$, and related $\beta$ to the structure of the empirical NTK, which becomes increasingly diagonal-dominant as gradients localize. These results turn $\beta$ into a simple but effective knob for controlling both stability and locality in implicit neural representations.

Empirically, HOSC acts as a drop-in replacement for sine activations. 
The gains are strongest on 1D audio, where high carrier frequencies are beneficial over SIREN and FINER. On images and videos HOSC behaves like a mild regularizer yielding smaller but robust improvements. On NeRF it matches frequency-tuned baselines while remaining stable at more aggressive hyper-parameters. Overall, our results suggest that explicitly designing activations for controllable Lipschitz constants and localized gradients is complementary to frequency-based INR architectures. As future work, we aim to develop activations and architectures that achieve an even fuller decoupling between frequency content and gradient control, for example via learnable carriers or spatially adaptive gating in high-dimensional neural fields.

{
    \small
    \bibliographystyle{ieeenat_fullname}    
    \bibliography{main}
    
}

\clearpage
\onecolumn

\maketitlesupplementary
\appendix

\section*{Appendix Overview}
This appendix first collects the mathematical details for HOSC activations, including formal statements and proofs for the results in Sec.~3. The subsequent sections provide practical information for the experiments: training and implementation details, ablations of the parameter $\beta$, and extended quantitative tables and qualitative visualizations for all image, audio, video, NeRF, and SDF experiments, as well as an additional hash-grid experiment. 
Qualitative results are available in an HTML experiment suite via the entry page (\url{\weblink}). 
\\[1pt]

\section{Mathematical Details for HOSC}
This part provides formal statements and proofs supporting the mathematical claims in Sec. 3. 

\subsection*{Notation,  assumptions, and terminology}
Let \(\beta>0\) and \(\omega_0>0\). Define the scalar activation
\[
\phi_{\beta,\omega_0}(x)=\tanh\!\big(\beta\,\sin(\omega_0 x)\big).
\]
For layers use \(z=Wx+b\) and apply \(\phi\) elementwise. Matrix norms are spectral (\(\|\cdot\|_2\)); Jacobian/operator norms are the induced \(\ell_2\)-norms. ``Activation‑level Lipschitz'' is w.r.t. preactivation \(z\). 
\noindent
\textbf{Terminology note: }
``Decouple'' in the main text means introducing an independent knob \(\beta\) for gradient scale while \(\omega_0\) sets the carrier frequency; the activation‑level Lipschitz equals \(\beta\omega_0\) and depends on both.\\[1pt]

\subsection{Activation calculus and Lipschitz bound}

\paragraph{Lemma A.1.1 (Derivative).}
For \(f(x)=\phi_{\beta,\omega_0}(x)\),
\[
f'(x)=\beta\,\omega_0\,\cos(\omega_0 x)\,\operatorname{sech}^2\!\big(\beta\,\sin(\omega_0 x)\big).
\]
\textbf{Proof.} Chain rule with \(u(x)=\beta\sin(\omega_0 x)\) and \(\frac{d}{dx}\tanh u=\operatorname{sech}^2 u\cdot u'\). \(\square\)

\paragraph{Lemma A.1.2 (Tight activation‑level Lipschitz).}
\[
\displaystyle \sup_{x\in\mathbb{R}}|f'(x)|=\beta\,\omega_0.
\]
\noindent
\textbf{Proof.} From Lemma A.1.1, \(|f'|\le \beta\omega_0\) since \(|\cos|\le 1\) and \(0<\operatorname{sech}^2\le 1\). Equality at \(x=k\pi/\omega_0\) where \(\sin=0\) and \(|\cos|=1\). \(\square\)

\paragraph{Corollary A.1.3 (Basic properties).}
\(\phi_{\beta,\omega_0}\) is \(2\pi/\omega_0\)-periodic, odd, bounded in \([-1,1]\), and \(C^\infty\).\\[1pt]

\subsection{Layer and network bounds}

\paragraph{Proposition A.2.1 (Layer Jacobian bound).}

For \(y=\phi_{\beta,\omega_0}(Wx+b)\),
\[
\|\nabla_x y\|_2 \;\le\; \beta\,\omega_0\,\|W\|_2 \quad \text{for all }x.
\]

\noindent
\textbf{Proof.} The Jacobian factors as \(J=D(z)\,W\) with \(D(z)=\mathrm{diag}(f'(z_i))\). By Lemma A.1.2, \(\|D(z)\|_2\le \beta\omega_0\). Submultiplicativity gives \(\|J\|_2\le \beta\omega_0\|W\|_2\). \(\square\)

\paragraph{Proposition A.2.2 (Network Lipschitz \emph{upper} bound).}
For an \(L\)-layer MLP \(y_L\) with \(y_k=\phi_{\beta,\omega_0}(W_k y_{k-1}+b_k)\),
\[
\|\nabla_x y_L(x)\|_2 \;\le\; \prod_{k=1}^L \big(\beta\,\omega_0\,\|W_k\|_2\big),\quad \forall x.
\]

\noindent
\textbf{Proof.} Apply Proposition A.2.1 to each layer and submultiplicativity to the Jacobian product. \(\square\)\\[1pt]

\subsection{Asymptotics in \texorpdfstring{\(\beta\)}{beta}}

\paragraph{Proposition A.3.1 (Small‑\(\beta\) expansion).}
For \(\beta\in(0,\beta_0]\),
\[
\phi_{\beta,\omega_0}(x)=\beta\sin(\omega_0 x)-\frac{\beta^3}{3}\sin^3(\omega_0 x)+O(\beta^5)
\]
uniformly in \(x\).

\noindent
\textbf{Proof.} Taylor series \(\tanh s=s-s^3/3+O(s^5)\) with \(s=\beta\sin(\omega_0 x)\) and \(|\sin|\le 1\). \(\square\)

\paragraph{Proposition A.3.2 (Large‑\(\beta\) limit).}
If \(\sin(\omega_0 x)\neq 0\), then \(\phi_{\beta,\omega_0}(x)\to \mathrm{sign}(\sin(\omega_0 x))\) as \(\beta\to\infty\); convergence holds pointwise a.e. and in \(L^p_{\mathrm{loc}}\) for any finite \(p\).

\noindent
\textbf{Proof.} \(\tanh y\to \mathrm{sign}(y)\) as \(|y|\to\infty\). Take \(y=\beta\sin(\omega_0 x)\). Dominated convergence yields local \(L^p\) convergence on compact sets away from zeros; boundedness by 1 provides domination. \(\square\)\\[1pt]

\subsection{Gating region measure}\label{sec:appx:asymptotics}
This section addresses Section~\ref{sec:asymp_behavior} in the paper. 
Let
\begin{align*}    
g_\beta(x)
&=\beta\,\omega_0\,|\cos(\omega_0 x)|\,\operatorname{sech}^2\!\bigl(\beta\,|\sin(\omega_0 x)|\bigr),
\\
u&=\omega_0 x\in[0,2\pi].
\end{align*}

\paragraph{Lemma A.4.1 (Per-period active set under proportional thresholds).}

Fix $\kappa\in(0,1)$ and define the proportional threshold
\[
\tau_\beta=\kappa\,\beta\,\omega_0.
\]
Over one period $I=[0,2\pi/\omega_0]$, consider the superlevel set
\[
S_\beta=\{\,x\in I:\ g_\beta(x)\ge \tau_\beta\,\}.
\]
Then
\[
|S_\beta|
\;\le\;
\frac{4}{\omega_0}\,
\arcsin\!\Bigl(
\min\!\Bigl\{
1,\,
\frac{\operatorname{arcosh}(1/\sqrt{\kappa})}{\beta}
\Bigr\}
\Bigr),
\]
and in particular $|S_\beta| = O(1/\beta)$ as $\beta\to\infty$.

\noindent
\textbf{Proof.}
Write $u=\omega_0 x$.  The inequality
\[
g_\beta(x)
=\beta\omega_0\,|\cos u|\,
\operatorname{sech}^2\!\bigl(\beta|\sin u|\bigr)
\;\ge\;
\kappa\beta\omega_0
\]
\emph{implies that} $\operatorname{sech}^2(\beta|\sin u|)\ge\kappa$, since $|\cos u|\le 1$.
Thus a \emph{necessary} condition for membership in $S_\beta$ is
\[
\operatorname{sech}^2(\beta|\sin u|)\ge\kappa
\,\Longleftrightarrow\,
\beta|\sin u|
\le
c_\kappa
=
\operatorname{arcosh}\!\bigl(1/\sqrt{\kappa}\bigr).
\]
Hence
\[
S_\beta
\;\subset\;
\Bigl\{\,x\in I:\ |\sin(\omega_0 x)|\le c_\kappa/\beta\,\Bigr\}.
\] 
Over $u\in[0,2\pi]$, the set $\{\,u:\ |\sin u|\le\varepsilon\,\}$ consists of three intervals
(near $0$, $\pi$, and $2\pi$) with total length $4\arcsin(\varepsilon)$.
Mapping back via $x=u/\omega_0$ gives
\[
|S_\beta|
\;\le\;
\frac{4}{\omega_0}\,
\arcsin\!\Bigl(\min\{1,\,c_\kappa/\beta\}\Bigr).
\]
As $\beta\to\infty$, we have $c_\kappa/\beta\to 0$ and therefore
\[
\arcsin\!\Bigl(\min\{1,\,c_\kappa/\beta\}\Bigr)
=
\arcsin(c_\kappa/\beta)
\sim c_\kappa/\beta,
\]
which implies $|S_\beta| = O(1/\beta). \;\square$ \\[1pt]

\subsection{Tightness points}

\paragraph{Lemma A.5.1 (Where the maximum slope is attained).}
\(|f'(x)|=\beta\omega_0\) iff \(x=k\pi/\omega_0\), \(k\in\mathbb{Z}\).
\noindent 
\textbf{Proof.} Simultaneously require \(|\cos(\omega_0 x)|=1\) and \(\sin(\omega_0 x)=0\). \(\square\)\\[1pt]

\subsection{Multivariate extension}

For elementwise activations, \(z=Wx+b\in\mathbb{R}^m\) and \(y=\phi_{\beta,\omega_0}(z)\) applied elementwise, the diagonal Jacobian satisfies \(\|\nabla_z y\|_2\le \beta\omega_0\) (Lemma A.1.2), yielding Proposition A.2.1 by composition with \(W\).
\\[10pt]

\section{Practical Experiments Details}

All experiments were conducted under the same experimental regime described in the paper. At first an ablation study on the HOSC $\beta$ parameter is conducted. After that a comparison with known methods is computed. For each experiment we present the implementation and experimental details along with additional results that have been omitted in the paper and left out for the supplemental.

\noindent
\textbf{Qualitative Results.}
For convenient inspection of our results we provide an HTML experiment suite in the supplemental material. The entry page (\url{\weblink}) links to per-experiment pages for image, audio, video, NeRF, and SDF fitting. Each page shows the same methods and test cases as the corresponding tables/figures and offers interactive browsing (e.g., image and video sliders, audio play–pause controls) with the best methods highlighted as in the quantitative results.

\noindent
\textbf{Source Code.}
We will publicly release the full source code for all experiments presented in this paper, including training scripts, evaluation pipelines, and the HTML experiment suite, to facilitate reproducibility and further research.

\noindent
\textbf{Method Comparison Setup.} We compare HOSC with methods like PEMLP (FF) \cite{tancik2020ffn}, Gauss \cite{ramasinghe2022periodicity}, WIRE \cite{saragadam2023wire}, \cite{sitzmann2020siren} and FINER \cite{liu2024finer}.
For each baseline method, the hyperparameters were configured according to the specifications provided in their original publications or as implemented in their respective reference codebases. Specifically, for PEMLP (FF), the Fourier Feature mapping utilized a total of 10 frequencies. The Gauss's scale parameter was set to 30. The WIRE configuration involved setting both the initial ($\omega_0$) and hidden ($\omega_h$) $\omega$ parameters to 20, while the scale factor was set to 10. A common setup was employed for SIREN, where the initial $\omega_0$ was set to 30 and all subsequent layer-wise $\omega$ parameters were uniformly set to 1. Finally, the specific parameter configuration for FINER was adopted directly from the setup described in its official implementation.\\[1pt]

\begin{table*}[!t]
\centering
\small
\setlength\tabcolsep{4pt}
\caption{\textbf{Audio HOSC $\beta$-ablation}. Showcased for Bach and Counting examples.}
\label{tab:audio_ablation-appx}

\vspace{-1.2em}
\begin{tabular*}{\textwidth}{@{\extracolsep{\fill}}ccccccccc}
\toprule
& \multicolumn{8}{c}{\textbf{HOSC } $\boldsymbol{\beta}$} \\
\cmidrule(lr){2-9}
Example & 
\multirow{2}*{0.1} &
\multirow{2}*{0.2} &
\multirow{2}*{0.5} &
\multirow{2}*{0.8} &
\multirow{2}*{1.0} &
\multirow{2}*{2.0} &
\multirow{2}*{4.0} &
\multirow{2}*{5.0} \\
MSE $\downarrow$ &  &  &  &  &  &  \\

\midrule
Bach & 6.6636e-04 & 2.2016e-04 & 1.7386e-05 & 1.0024e-05 & 9.8558e-06 & \cellcolor{yellow}2.0293e-06 & \cellcolor{red}7.5893e-07 & \cellcolor{orange}1.3194e-06 \\
Counting & 9.8662e-04 & 5.2919e-04 & 4.1339e-04 & 4.0398e-04 & 4.3283e-04 & 2.5501e-04 & \cellcolor{yellow}6.8511e-05 & \cellcolor{red}3.6126e-05 \\

\midrule
& \multirow{2}*{6.0} & \multirow{2}*{8.0} & \multirow{2}*{10.0} & \multirow{2}*{12.0} & \multirow{2}*{14.0} & \multirow{2}*{16.0} & \multirow{2}*{18.0} & \multirow{2}*{20.0} \\
 &  &  &  &  &  &  &  &  \\
\midrule
Bach & \cellcolor{orange}2.3844e-06 & 2.1383e-05 & 9.4639e-05 & 4.3760e-04 & 2.3118e-02 & 2.4166e-02 & 2.4180e-02 & 2.4184e-02 \\
Counting & 9.3554e-05 & 1.7069e-04 & 8.9352e-04 & 6.7790e-03 & 7.2318e-03 & 7.0961e-03 & 7.3637e-03 & 7.3835e-03 \\
\bottomrule
\end{tabular*}
\end{table*}

\begin{table*}[!t]
\centering
\small
\setlength\tabcolsep{10.7pt}
\caption{\textbf{Seed Independence}. Seed independence evaluation of HOSC and related methods.}
\label{tab:audio_seed}

\vspace{-1.2em}
\begin{tabular*}{\textwidth}{@{\extracolsep{\fill}}ccccccc}
\toprule
& \multicolumn{2}{c}{\textbf{SIREN}} & \multicolumn{2}{c}{\textbf{FINER}} & \multicolumn{2}{c}{\textbf{HOSC}} \\
\cmidrule(lr){2-3} \cmidrule(lr){4-5} \cmidrule(lr){6-7}
Example & 
\multirow{2}*{Mean} &
\multirow{2}*{Std} &
\multirow{2}*{Mean} &
\multirow{2}*{Std} &
\multirow{2}*{Mean} &
\multirow{2}*{Std}  \\
MSE $\downarrow$ &  &  &  &  &  &  \\

\midrule
Bach & \cellcolor{orange}5.6905e-06 & \cellcolor{orange}3.9998e-07 & \cellcolor{yellow}1.2704e-3 & \cellcolor{yellow}4.0961e-03 & \cellcolor{red}7.1032e-07 & \cellcolor{red}5.8887e-08 \\
Counting & \cellcolor{orange}3.853e-04 & \cellcolor{orange}4.5750e-06 & \cellcolor{yellow}2.8612e-03 & \cellcolor{yellow}1.6933e-03 & \cellcolor{red}4.4506e-05 & \cellcolor{red}5.6223e-06 \\
\bottomrule
\end{tabular*}
\end{table*}

\subsection{Image Fitting}
\noindent
\textbf{Data.}
We utilize a 16 example subset from the testing set of the DIV2K Dataset \cite{Agustsson_2017_CVPR_Workshops}. Each image is of $512 \times 512$ resolution. Both the value range and input pixel coordinates are scaled to $[-1,1]$. No additional preprocessing is done. We later rescale both ranges for inference and results collection.

\noindent
\textbf{Network \& Hyperparameters.} We use an MLP with 3 hidden layers, of 256 width, for all experiments on fitting images. We train for 5000 epochs, at each iteration we fit on every pixel of the GT image, batch size is equal to the number of pixels in the image. We use the Adam optimizer with a learning rate specific to each method, consistent throughout each trained sample. We utilize activation specific recommended learning rates. For FINER, SIREN and HOSC we use a LR$=5e-04$, for Gauss and WIRE we use a LR$=5e-03$ and for PEMLP we use a LR$=1e-3$. We conducted experiments on consistent learning rates on all methods and found out that the mentioned rates achieve a better reconstruction quality.

\noindent
\textbf{Hardware \& Runtime.} The networks are trained using a single NVIDIA A100 GPU with 40GB of memory. We train for 5,000 epochs, requiring approximately 3 minutes to fit and evaluate a single image. For the ablation study we consider 256 configurations which all ammount to 16hrs of training. The comparison study took 4.8hrs of training. \\[1pt]

\begin{table*}[t]
\centering
\small
\setlength{\tabcolsep}{10.0pt}
\renewcommand{\arraystretch}{1.2}
\caption{\textbf{Quantitative comparisons}. Averaged over eight NeRF Synthetic \cite{mildenhall2020nerf} samples. }
\vspace{-1.2em}
\begin{tabular*}{\textwidth}{@{\extracolsep{\fill}}cccccccc}
\toprule
&
Stat. & 
{PEMLP} & {Gauss} &
{WIRE} & {SIREN} &
{FINER} & {HOSC} \\
\midrule

\multirow{2}*{\rotatebox{90}{PSNR}} 
& Mean.
& 23.9840
& 26.1985
& 29.2248
& \cellcolor{yellow}29.7097
& \cellcolor{red}29.8378
& \cellcolor{orange}29.7321
\\
& Std.
& 3.4295
& 3.7701
& 4.9856
& 4.6224
& 4.5956
& 4.6157 
\\
\cmidrule(lr){2-8}

\multirow{2}*{\rotatebox{90}{SSIM}} &
Mean.
& 0.8811
& 0.8877
& 0.9303
& \cellcolor{yellow}0.9371
& \cellcolor{orange}0.9371
& \cellcolor{red}0.9388 \\
& Std.
& 0.0779
& 0.0852
& 0.0772
& 0.0653
& 0.0660
& 0.0622\\
\cmidrule(lr){2-8}

\multirow{2}*{\rotatebox{90}{LPIPS}} &
Mean.
& 0.1078
& 0.0856
& 0.0485
& \cellcolor{red}0.0355
& \cellcolor{yellow}0.0352
& \cellcolor{orange}0.0353 \\ 
& Std.
& 0.0681
& 0.0617
& 0.0575
& 0.0302
& 0.0313
& 0.0277\\

\bottomrule
\end{tabular*}
\label{tab:nerf_ablation-appx}
\end{table*}

\subsection{Audio Fitting}
\noindent
\textbf{Data.} For learning audio signals we use two samples; the first 7 seconds from Bach's Cello Suite No. 1, representing music data and a stock audio of a male actor counting from 0 to 9, representing human speech data. Audio signals are taken from waveform files sampled at 44100 samples per second. Both value and coordinates are scaled to the range of $[-1,1]$. We use all sample points as training data. We provide additional results for the Counting sample in the HTML experiments suite (\url{\weblink}) and an extended $\beta$ parameter ablation in Tables~\ref{tab:audio_ablation} and~\ref{tab:audio_ablation-appx}. 

\noindent
\textbf{Network \& Hyperparameters.} We use an MLP with 3 hidden layers, of 256 width, for all experiments on fitting audio. We train for 5000 epochs, at each iteration we fit on the whole audio wave, batch size is equal to the number sample tick. We use the Adam optimizer with a learning rate specific to each method, exactly the same as in the setup for image experiments.

\noindent
\textbf{Hardware \& Runtime.} The networks are trained using a single NVIDIA A100 GPU with 40GB of memory. We train for 5,000 epochs, requiring approximately 2 minutes to fit and evaluate a single audio waveform. For the ablation study we consider 90 configurations which all ammount to 30 minutes of training. The comparison study took 1hrs of training. 

\noindent
\textbf{Seed Independence.} We evaluate the impact of seed choice on the quality of result by training audio reconstruction models on a finite set of chosen seeds: $\{0, 2, 4, 8, 12, 21, 42, 1337, 3333, 3407, 2025\}$. We see a very small dependency on seed for both SIREN and HOSC, while FINER due to its inherent reliance on initialization is considerably influence by the seed change, see Tab. \ref{tab:audio_seed}.
\\[1pt]

\subsection{Video Fitting}

\noindent
\textbf{Data.} For learning videos we use a single video of a Cat which contains 300 frames of $512\times512$ resolution.
The value range is scaled to $[-1,1]$. Full Quantitative results are available in the paper.

\noindent
\textbf{Network \& Hyperparameters.} We use an MLP with 5 hidden layers, of 1024 width, for all experiments on fitting video. We train for 100 000 epochs, at each iteration we fit on the whole video at once, batch size is equal to the whole video size. We use the Adam optimizer with a learning rate of $1e-04$ consistent across all methods. We found that method specific learning rates used for image and audio reconstruction, achieved inconclusive results. While on method specific learning rates PEMLP achieved a PSNR of $37.12$ outperforming SIREN, the other methods struggled to achieve a PSNR of over $14.00$. In the paper we present results for consistent learning rate comparisons only.

\noindent
\textbf{Hardware \& Runtime.} The networks are trained using a single NVIDIA H100 GPU with 80GB of memory. We train for 100 000 epochs, requiring approximately 10hrs to fit and evaluate a single video. For the ablation study we consider 5 configurations which all ammount to 50hrs of training. The comparison study took 60hrs of training.\\[1pt]

\subsection{Neural Radiance Fields (NeRF)}

\noindent
\textbf{Data.} In Neural Radience Fields experiments  we use the NeRF Synthetic Dataset \cite{mildenhall2020nerf} consisting of 8 examples. We utilized the experimental setting described in WIRE where only 25 images are used for training, and each image is downsampled to a resolution of $200\times200$, 4$\times$ downsampling from the original $800\times800$.

\noindent
\textbf{Network \& Hyperparameters.} We use an MLP with 4 layers, of 182 width, for all NeRF experiments. We train for 37500 iterations. We use the Adam optimizer with a scheduled learning rate starting from 2e-4 and finishing at 1e-15, consistent across all methods. We don't consider method specific learning rates, due to the learning rate schedulers influence. 

\noindent
\textbf{Hardware \& Runtime.} The networks are trained using a single NVIDIA A100 GPU with 40GB of memory. We train for 30 000 iterations, requiring approximately 30 minutes to fit and evaluate a single NeRF. For the ablation study we consider 56 configurations which all ammount to approximately 30hrs of training. The comparison study took 15hrs of training.\\[1pt]

\begin{table*}[t]
\centering
\footnotesize
\setlength\tabcolsep{4.3pt}
\caption{\textbf{Signed Distance Field HOSC $\beta$-ablation}. Experiment on the \texttt{Arm\-a\-dil\-lo} shape.}

\vspace{-1.2em}
\begin{tabular}{cccccccccccc}

\toprule
& \multicolumn{11}{c}{\textbf{HOSC } $\boldsymbol{\beta}$} \\
\cmidrule(lr){2-12}
Metric & 
0.1 &
0.3 &
0.5 &
0.8 &
0.9 &
1.0 &
3.0 &
5.0 &
8.0 &
12.0 &
16.0 \\

\midrule
Chamfer &  
5.280e-06 &
\cellcolor{yellow}4.814e-06 &
5.638e-06 &
\cellcolor{orange}4.447e-06 &
5.581e-06 &
\cellcolor{red}4.324e-06 &
5.676e-06 &
6.057e-06 &
5.877e-06 &
5.293e-06 &
7.673e-06 \\

IoU & 
0.9774 &
0.9778 &
0.9728 &
\cellcolor{orange}0.9817 &
0.9762 &
\cellcolor{red}0.9821 &
0.9738 &
0.9729 &
0.9734 &
\cellcolor{yellow}0.9808 &
0.9670 \\

\bottomrule
\end{tabular}
\label{tab:sdf_beta_abl-appx}
\end{table*}

\begin{table*}[t]
\setlength\tabcolsep{8.5pt}
\small
\centering
\caption{\textbf{Quantitative comparisons on SDF reconstruction}. HOSC $\beta=\{0.8,0.9,1.0\}$ }
\vspace{-1.2em}
\begin{tabular}{ccccccccc}
\toprule
& \multirow{2}*{Samples}
& \multirow{2}*{Gauss} & \multirow{2}*{WIRE} & \multirow{2}*{SIREN} & \multirow{2}*{FINER} 
& {HOSC} & {HOSC} & {HOSC} \\
& & & & & & {$\beta=0.8$} & {$\beta=0.9$} & {$\beta=1.0$} \\

\midrule
\multirow{4}*{\rotatebox{90}{Chamfer $\downarrow$}} 
& {Armadillo} & 2.2065e-05 & \cellcolor{yellow}4.3975e-06 & 4.8958e-06 & \cellcolor{red}4.3178e-06 & 4.4470e-06 & 5.5813e-06 & \cellcolor{orange}4.3237e-06 \\

& {Dragon}  & 1.9011e-05 & \cellcolor{red}2.5875e-06 & \cellcolor{yellow}3.1763e-06 & \cellcolor{orange}3.0937e-06 & 3.3799e-06 & 3.9059e-06 & 3.7843e-06 \\

& {Thai}  & 3.7762e-05 & \cellcolor{red}2.8232e-05 & 3.1437e-05 & \cellcolor{orange}2.9137e-05 & \cellcolor{yellow}3.0911e-05 & 3.2658e-05 & 3.1313e-05\\

& {Lucy}  & 2.5397e-05 & \cellcolor{red}2.1952e-05 & \cellcolor{yellow}2.2656e-05 & \cellcolor{orange}2.2189e-05 & 2.2966e-05 & 2.3785e-05 & 2.2929e-05 \\ 

\midrule
\multirow{4}*{\rotatebox{90}{IoU $\uparrow$}} 
& {Armadillo}  & 0.9766 & 0.9799 & 0.9775 & \cellcolor{yellow}0.9809 & \cellcolor{orange}0.9817 & 0.9762 & \cellcolor{red}0.9821 \\

& {Dragon} & 0.9506 & \cellcolor{red}0.9617 & \cellcolor{yellow}0.9584 & \cellcolor{orange}0.9610 & 0.9546 & 0.9508 & 0.9498 \\

& {Thai}  & 0.8310 & \cellcolor{yellow}0.8328 & \cellcolor{orange}0.8338 & 0.8297 & \cellcolor{red}0.8366 & 0.8326 & 0.8229 \\

& {Lucy}  & 0.8625 & 0.8719 & \cellcolor{red}0.8731 & 0.8692 & \cellcolor{orange}0.8727 & 0.8675 & \cellcolor{yellow}0.8724 \\

\bottomrule
\end{tabular}
\label{tab:sdf_comp}
\end{table*}

\subsection{Signed Distance Fields}

\noindent
\textbf{Data.} In Signed Distance Fields experiments we use samples from the Stanford 3D Scanning Repository \cite{standord-3D-scanning}, the Armadillo, Dragon, Thai Statue and Lucy. Model vertices are normalized to fit inside the unit cube. 

\noindent
\textbf{Network \& Hyperparameters.} We use an MLP with 3 layers, of 256 width, for all SDF experiments. We use the Adam optimizer with a scheduled learning of $1e-04$, consistent across all methods. The coarse-to-fine loss function \cite{lindell2022bacon} is used. Training takes 200 000 iterations. In each iteration 10 000 points are randomly sampled and fed to the model. For visualization $512^3$ grid is extracted.

\noindent
\textbf{Hardware \& Runtime.} The networks are trained using a single NVIDIA A100 GPU with 40GB of memory. We train for 200 000 iterations, requiring approximately 3hrs to fit and evaluate a single SDF. For the ablation study we consider 11 configurations which all ammount to approximately 30hrs of training. The comparison study took 24hrs of training.\\[1pt]

\subsection{Orthogonality to encoding methods} 

\begin{figure*}
    \centering

    \setlength{\SpaceCaptionToRow}{-0.0em} 
    \setlength{\SpaceMetricToNextRow}{-0.0em} 
    \setlength{\SpaceImageToMetric}{-1.4em} 

    \vspace{\SpaceCaptionToRow}

    \def\subfigwidth{0.198\textwidth}
    
    \setlength{\VerticalLabelWidth}{0.6cm} 
    
    \setlength{\VerticalLabelHeight}{2.2cm} 
    \setlength{\VerticalLabelHeightDiff}{1.8cm} 

    \begin{tabular}{@{}c@{ }c@{}}
        \mbox{\hspace{\VerticalLabelWidth}} & %
        \resizebox{0.95\textwidth}{!}{%
        \begin{minipage}[t]{\textwidth}
            \centering
            \begin{minipage}{\subfigwidth}
                \centering\textbf{INGP}
            \end{minipage}%
            \hfill
            \begin{minipage}{\subfigwidth}
                \centering\textbf{PEMLP}
            \end{minipage}%
            \hfill
            \begin{minipage}{\subfigwidth}
                \centering\textbf{SIREN}
            \end{minipage}%
            \hfill
            \begin{minipage}{\subfigwidth}
                \centering\textbf{HOSC}
            \end{minipage}%
            \hfill
            \begin{minipage}{\subfigwidth}
                \centering\textbf{GT}
            \end{minipage}%
        \end{minipage}%
        }%
        \\ 

        \rotatebox{90}{\parbox{\VerticalLabelHeightDiff}{\centering\textbf{TCNN}}} &
        \resizebox{0.95\textwidth}{!}{%
        \begin{minipage}[t]{\textwidth}
            \begin{subfigure}[t]{\subfigwidth}
                \centering
                \includegraphics[width=\textwidth]{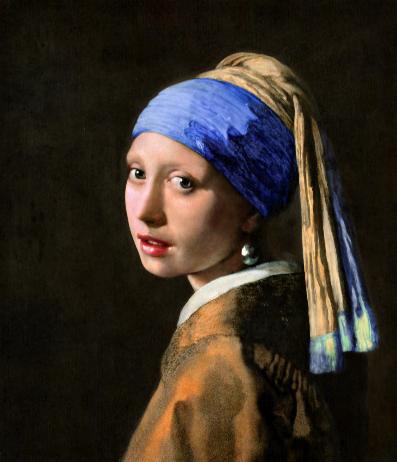}
                \vspace{\SpaceImageToMetric}
                \caption*{\textbf{25.1861}}
            \end{subfigure}%
            \hfill
            \begin{subfigure}[t]{\subfigwidth}
                \centering
                \includegraphics[width=\textwidth]{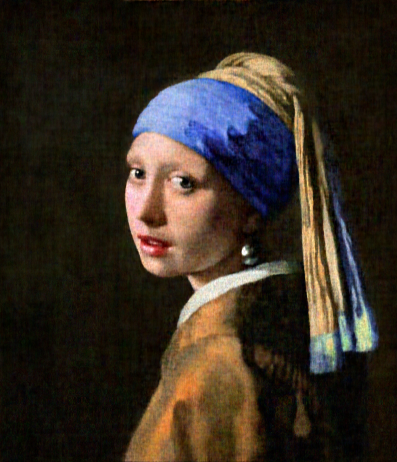}
                \vspace{\SpaceImageToMetric}
                \caption*{22.5088}
            \end{subfigure}%
            \hfill
            \begin{subfigure}[t]{\subfigwidth}
                \centering
                \includegraphics[width=\textwidth]{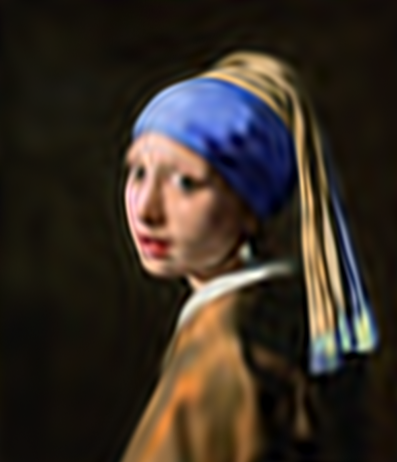}
                \vspace{\SpaceImageToMetric}
                \caption*{21.6710}
            \end{subfigure}%
            \hfill
            \begin{subfigure}[t]{\subfigwidth}
                \centering
                \includegraphics[width=\textwidth]{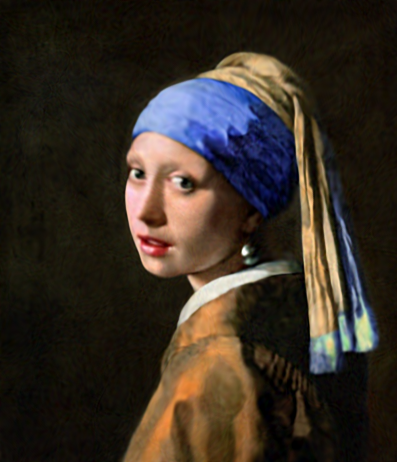}
                \vspace{\SpaceImageToMetric}
                \caption*{22.7124}
            \end{subfigure}%
            \hfill
            \begin{subfigure}[t]{\subfigwidth}
                \centering
                \includegraphics[width=\textwidth]{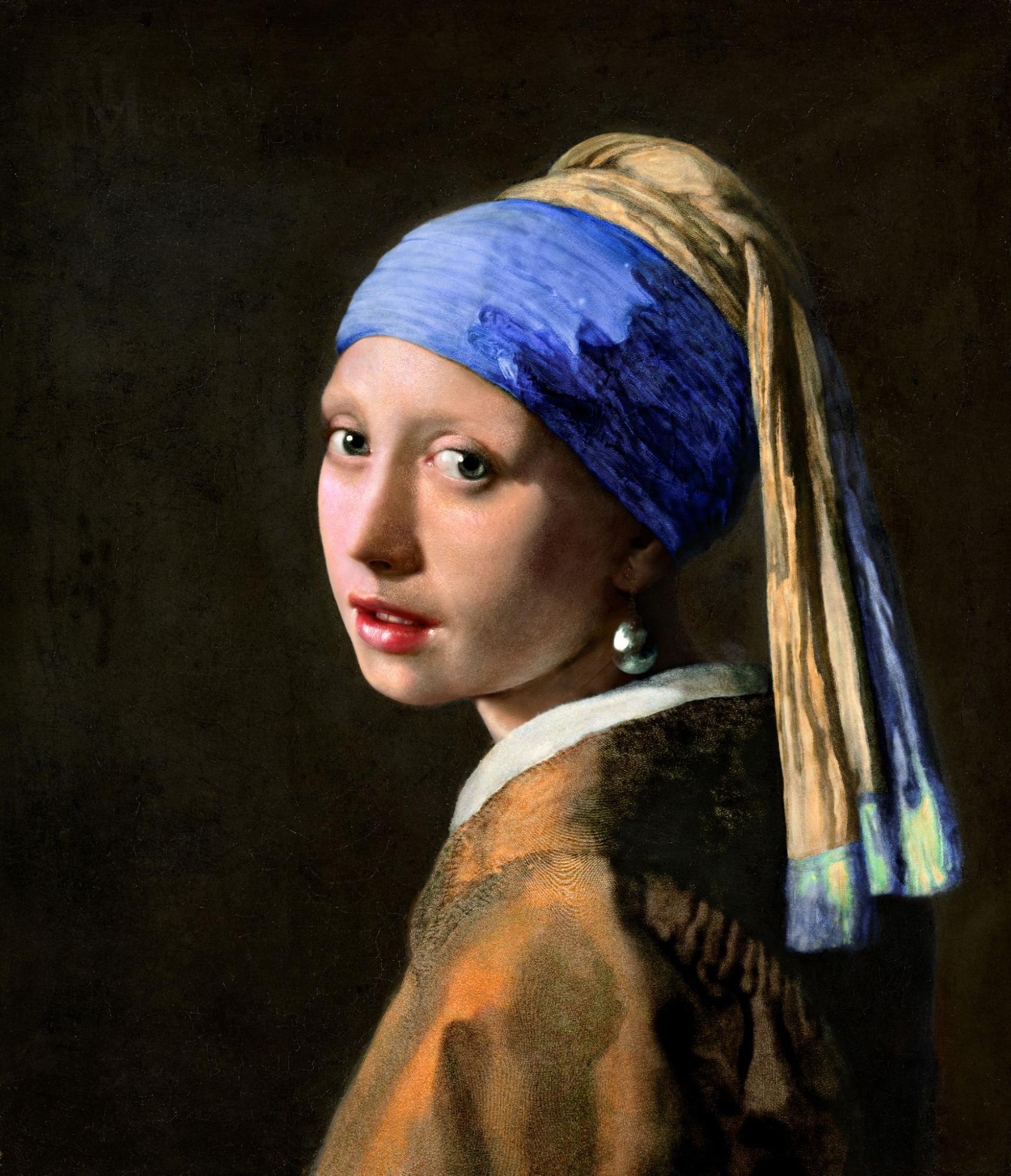}
                \vspace{\SpaceImageToMetric}
                \caption*{}
            \end{subfigure}%
            \hfill
        \end{minipage}%
        }%
        \\ 
        \noalign{\vspace{\SpaceMetricToNextRow}}

        \rotatebox{90}{\parbox{\VerticalLabelHeightDiff}{\centering\textbf{PyTorch}}} &
        \resizebox{0.95\textwidth}{!}{%
        \begin{minipage}[t]{\textwidth}
            \begin{subfigure}[t]{\subfigwidth}
                \centering
                \vspace{\SpaceImageToMetric}
                \caption*{}
            \end{subfigure}%
            \hfill
            \begin{subfigure}[t]{\subfigwidth}
                \centering
                \includegraphics[width=\textwidth]{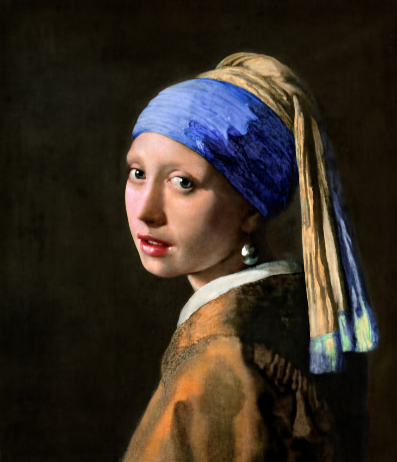}
                \vspace{\SpaceImageToMetric}
                \caption*{23.0268}
            \end{subfigure}%
            \hfill
            \begin{subfigure}[t]{\subfigwidth}
                \centering
                \includegraphics[width=\textwidth]{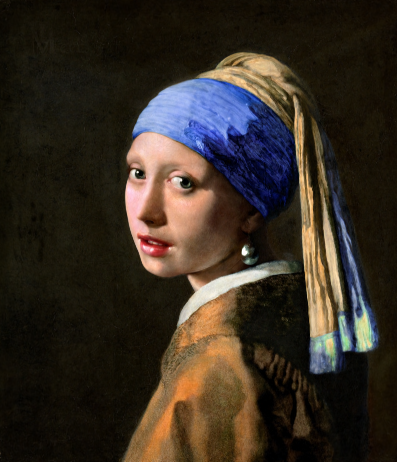}
                \vspace{\SpaceImageToMetric}
                \caption*{23.1933}
            \end{subfigure}%
            \hfill
            \begin{subfigure}[t]{\subfigwidth}
                \centering
                \includegraphics[width=\textwidth]{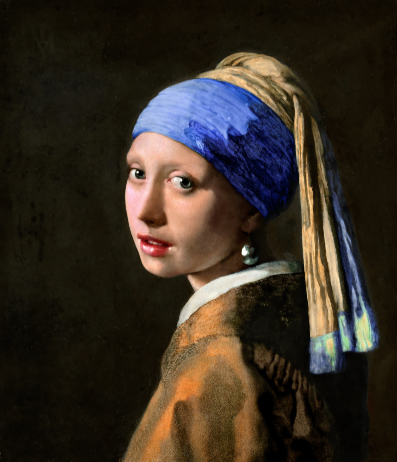}
                \vspace{\SpaceImageToMetric}
                \caption*{23.4463}
            \end{subfigure}%
            \hfill
            \begin{subfigure}[t]{\subfigwidth}
                \centering
                \includegraphics[width=\textwidth]{data/giga/tcnn/GT/girl.jpg}
                \vspace{\SpaceImageToMetric}
                \caption*{}
            \end{subfigure}%
            \hfill
        \end{minipage}%
        }%
        \\ 
    \end{tabular}
    \caption{Qualitative comparison on Gigapixel reconstruction. Each row reconstructs the GT data with column specified method. Quantitative (PSNR) results showcased per example below reconstructed images. HOSC $\beta=8.0$. for TCNN and $\beta=5.0$ for PyTorch. }
    \label{fig:giga_comparison_app}
\end{figure*}

To validate HOSC's scalability and address the relationship between activation design and spatial encoding methods, we implemented HOSC in both PyTorch and \textit{Tiny-Cuda-NN} \cite{tiny-cuda-nn} and trained on gigapixel-scale images. 
We compare HOSC against activation-based methods (SIREN, PEMLP) and Instant-NGP's HashGrid encoding \cite{mueller2022instant}, which represents the state-of-the-art for large-scale spatial data. Results are shown in Fig.~\ref{fig:giga_comparison_app} and Tables~ \ref{tab:sdf_beta_abl},~\ref{tab:giga_comp}.

\noindent
\textbf{Data.} For gigapixel image fitting experiments we use
the ``Girl With a Pearl Earring'' painting renovation ©Koorosh Orooj (CC BY-SA 4.0) at $8000\times9302$ resolution.

\begin{figure}[b]
    \centering
    \includegraphics[width=0.49\textwidth, trim={2.5cm 1.5cm 2.5cm 2cm}, clip]{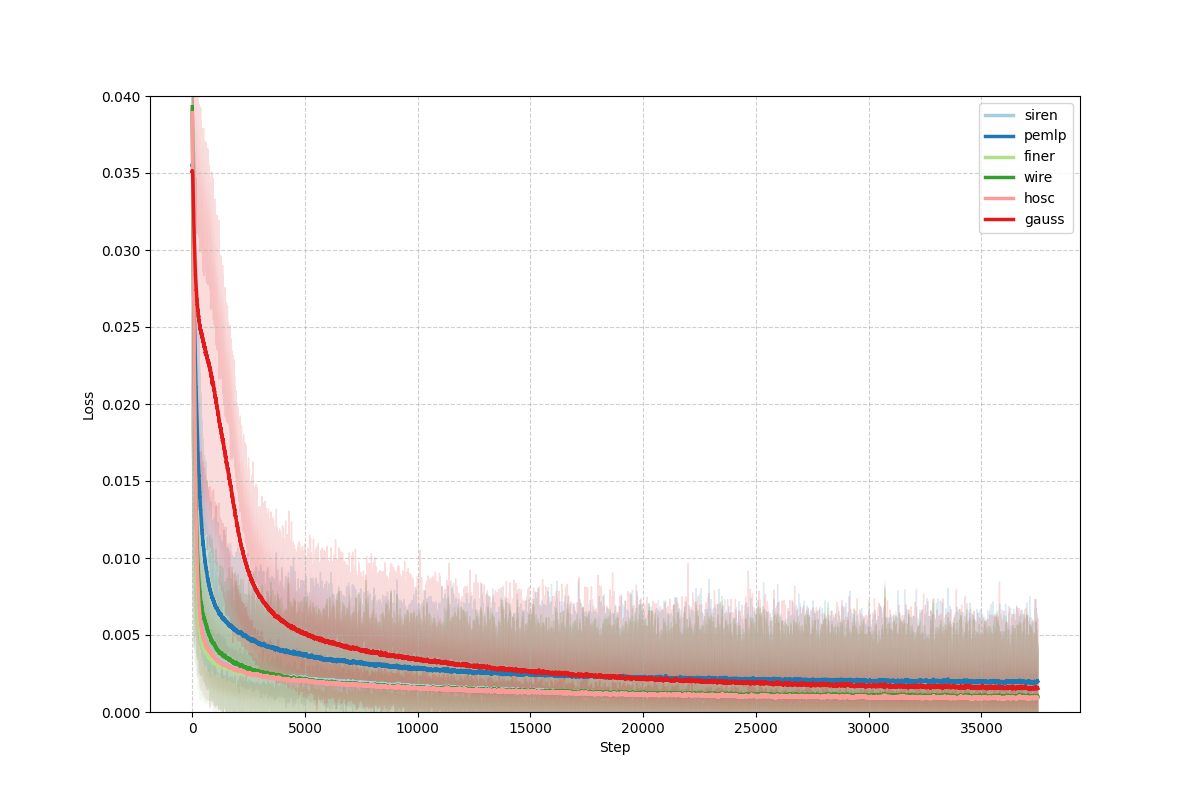} %
    \caption{\textbf{Method specific loss curves}. Average over all examples computed for NeRF experiments.}
    \label{fig:nerf_curves}
\end{figure}

\noindent
\textbf{Network \& Hyperparameters.} We use an MLP with 3 hidden layers of 256 width. We use the Adam optimizer with a learning rate of $1\times10^{-4}$, consistent across methods. We compare four methods: PEMLP, SIREN, HOSC, and the HashGrid encoding from Instant-NGP \cite{mueller2022instant}. Training is done for 1000 epochs with a batch size of 2,000,000 ($\approx$3\% of the entire image).

\noindent
\textbf{Hardware \& Runtime.} The networks are trained using a single NVIDIA A100 GPU with 40GB of memory. 
Training for 1000 epochs requires approximately 3 hours per configuration. 
The $\beta$-ablation study (Table~\ref{tab:sdf_beta_abl}) involves 10 configurations totaling approximately 30 hours. The comparison study (Table~\ref{tab:giga_comp}) required 21 hours total.

\noindent
\textbf{Results: HOSC vs activation baselines.} 
HOSC achieves \textbf{23.45 dB PSNR} in PyTorch and 22.71 dB in TCNN (Table~\ref{tab:giga_comp}), outperforming SIREN (23.19 dB / 21.67 dB) and PEMLP (23.03 dB / 22.51 dB) in both frameworks. The $\beta$-ablation (Table~\ref{tab:sdf_beta_abl}) demonstrates robustness across $\beta \in [0.5, 12.0]$, with PyTorch achieving peak performance at $\beta=5.0$ (23.45 dB) and TCNN at $\beta=8.0$ (22.71 dB). Visual results (Fig.~\ref{fig:giga_comparison_app}) show HOSC produces high-fidelity reconstructions competitive with SIREN and PEMLP across both implementations.

\noindent
\textbf{Results: HOSC vs spatial encoding (INGP).} 
Instant-NGP's HashGrid encoding achieves 25.19 dB (Table~\ref{tab:giga_comp}), approximately 2 dB higher than activation-based methods. This gap is expected: HashGrid employs a multi-resolution hash table that explicitly models spatial coherence, whereas HOSC (and all activation functions) operates on individual coordinates without 
spatial structure. Critically, HOSC and HashGrid are \textbf{orthogonal design choice}s—HashGrid is an input encoding, HOSC is an activation function—and could potentially be combined in future work.

\noindent
\textbf{Discussion.} Trade-offs and when to use HOSC.
While HashGrid excels on large-scale spatial data (images, NeRF), it requires domain-specific architectural modifications and additional memory for the hash table. HOSC provides a simpler, activation-based alternative that works across diverse domains without specialized encodings: the same activation achieves strong results on images (Sec.~4.1), audio (Sec.~4.2), video (Sec.~4.3), NeRF (Sec.~4.4), and SDFs (Sec.~4.5). For applications where architectural simplicity, multi-domain generality, or explicit gradient control via $\beta$ are priorities, HOSC offers a compelling alternative. The consistent performance across PyTorch and TCNN implementations (Table~\ref{tab:giga_comp}) further demonstrates HOSC's robustness and practical applicability.

\begin{table}[b]
\centering
\small
\setlength\tabcolsep{4.5pt}
\caption{\textbf{Gigapixel HOSC $\beta$-ablation}.}
\label{tab:sdf_beta_abl}
\vspace{-0.8em}

\begin{tabular*}{\columnwidth}{@{\extracolsep{\fill}}lccccc}
\toprule
& \multicolumn{5}{c}{\textbf{HOSC } $\boldsymbol{\beta}$} \\
\cmidrule(lr){2-6}
Framework & 0.5 & 1.0 & 5.0 & 8.0 & 12.0 \\
\midrule
TCNN    & 21.05 & 21.73 & \cellcolor{yellow}22.57 & \cellcolor{red}22.71 & \cellcolor{orange}22.69 \\
PyTorch & \cellcolor{orange}23.20 & \cellcolor{yellow}23.09 & \cellcolor{red}23.45 & 22.70 & 22.50 \\
\bottomrule
\end{tabular*}
\end{table}

\begin{table}[b]
\centering
\small
\setlength\tabcolsep{3pt}
\caption{\textbf{Quantitative comparison for Gigapixel reconstruction}.}
\label{tab:giga_comp}
\vspace{-0.8em}

\begin{tabular*}{\columnwidth}{@{\extracolsep{\fill}}lccccccc}
\toprule
& \multicolumn{4}{c}{\textbf{TCNN}} & \multicolumn{3}{c}{\textbf{PyTorch}}\\
\cmidrule(lr){2-5} \cmidrule(lr){6-8}
Metric & INGP & PEMLP & SIREN & HOSC & PEMLP & SIREN & HOSC \\
\midrule
PSNR & \cellcolor{red}25.19 & 22.51 & 21.67 & 22.71 & 23.03 & \cellcolor{yellow}23.19 & \cellcolor{orange}23.45 \\
\bottomrule
\end{tabular*}
\end{table}

\end{document}